%% file: arxiv.tex
\documentclass{article}

\PassOptionsToPackage{compress}{natbib}

\usepackage[preprint]{neurips_2025}

\usepackage[utf8]{inputenc} %
\usepackage[T1]{fontenc}    %
\usepackage{hyperref}       %
\hypersetup{
  colorlinks,
  linkcolor={red!50!black},
  citecolor={blue!50!black},
  urlcolor={blue!80!black}
}
\usepackage{url}            %
\usepackage{booktabs}       %
\usepackage{amsfonts}       %
\usepackage{nicefrac}       %
\usepackage{microtype}      %
\usepackage{xcolor}         %
\usepackage{makecell}

\usepackage{def}            %
\usepackage{algorithm}
\usepackage{algpseudocode}
\usepackage{enumitem}

\title{Efficient Controllable Diffusion \\ via Optimal Classifier Guidance}

\usepackage[nameinlink, capitalise]{cleveref}
\usepackage{wrapfig}

\usepackage{authblk}
\author[1]{\textbf{Owen Oertell}\textsuperscript{*}}
\author[1]{\textbf{Shikun Sun}\textsuperscript{*}}
\author[1]{\textbf{Yiding Chen}\textsuperscript{*}}
\author[1]{\protect\\\vspace{1mm}\textbf{Jin Peng Zhou}}
\author[2]{\textbf{Zhiyong Wang}}  %
\author[1]{\textbf{Wen Sun}}

\affil[1]{Cornell University}
\affil[2]{CUHK}

\begin{document}

\renewcommand\thefootnote{\fnsymbol{footnote}}
\footnotetext[1]{Equal contribution. Correspondence to \texttt{ojo2@cornell.edu}. }
\renewcommand{\thefootnote}{\arabic{footnote}}

\maketitle

\newcommand{\ALG}{SLCD}

\begin{abstract}
  The controllable generation of diffusion models aims to steer the model to generate samples that optimize some given objective functions. It is desirable for a variety of applications including image generation, molecule generation, and DNA/sequence generation. Reinforcement Learning (RL) based fine-tuning of the base model is a popular approach but it can overfit the reward function while requiring significant resources. We frame controllable generation as a problem of finding a distribution that optimizes a KL-regularized objective function. 
  We present \ALG{} -- Supervised Learning based Controllable Diffusion, which iteratively generates online data and trains a small classifier to guide the generation of the diffusion model. Similar to the standard classifier-guided diffusion, \ALG{}'s key computation primitive is classification and does not involve any complex concepts from RL or control. Via a reduction to no-regret online learning analysis, we show that under KL divergence, the output from \ALG{} provably converges to the optimal solution of the KL-regularized objective. Further, we empirically demonstrate that \ALG{} can generate high quality samples with nearly the same inference time as the base model in both image generation with continuous diffusion and biological sequence generation with discrete diffusion. Our code is available at \href{https://github.com/Owen-Oertell/slcd}{https://github.com/Owen-Oertell/slcd}. \looseness=-1
\end{abstract} 

\input{sec/introduction}

\input{sec/related}

\input{sec/preliminaries}

\input{sec/algorithm}

\input{sec/analysis}

\input{sec/experiments}

\input{sec/conclusion}
\input{sec/ack}

\bibliographystyle{ref}
\bibliography{ref}

\newpage
\appendix
\input{sec/appendix}

\end{document}

%% file: sec/introduction.tex
\section{Introduction}
\label{sec:introduction}

Diffusion models are an expressive class of generative models which are able to model complex data distributions \citep{song2021denoising,song2021scorebased}. Recent works have utilized diffusion models for a variety of modalities: images, audio, and molecules \citep{saharia2022photorealistic, ho2022imagen,uehara2024finetuning,hoogeboom2022equivariant}. However, modeling the distribution of data is often not enough for downstream tasks. Often, we want to generate data which satisfies a specific property, be that a prompt, a specific chemical property, or a specific structure.

Perhaps the simplest approach is classifier-guided diffusion where a classifier is trained using a pre-collected labeled dataset. The score of the classifier is used to guide the diffusion process to generate images that have high likelihood being classified under a given label. However this simple approach requires a given labeled dataset and is not directly applicable to the settings where the goal is to optimize a complicated objective function (we call it reward function hereafter). To optimize reward functions, Reinforcement Learning (RL) and stochastic optimal control based approaches have been studied \citep{black2024ddpo,oertell2024rlcm, domingoenrich2025adjoint, clark2024directly, fan2023DPOK, uehara2024finetuning, uehara2024feedback}.  These methods formulate the diffusion generation process as a continuous time Markov decision process and then apply RL optimizers such as proximal policy optimization (PPO) \citep{schulman2017proximal} or optimal control methods to optimize the given reward function. While these RL based approaches can optimize reward, they make the solution concept of fine-tuning diffusion model complicated: they often need to fine-tune the entire models, update the initial distribution or adjust the noise scheduler, and popular RL optimizers such as PPO are known to be notoriously instable, slow, and expensive to train. Diffusion model is known for its simplicity in training (i.e., a simple reduction to least square regression \citep{ho2020denoising}), applying RL or control based solutions on top it takes away the simplicity and beauty of the original diffusion model training. From a theoretical perspective, diffusion model convergence theory is constructed by a simple reduction to supervised learning (least square regression) \citep{chen2022sampling}. Once we model fine-tuning diffusion model as an RL problem, it is unlikely one can prove any meaningful optimality (e.g., global convergence, sample complexity) since modern RL theory is often limited to very restricted settings (e.g., small size Markov Decision Process) \citep{agarwal2019reinforcement}. 

\begin{figure}
    \centering
    \includegraphics[width=\linewidth]{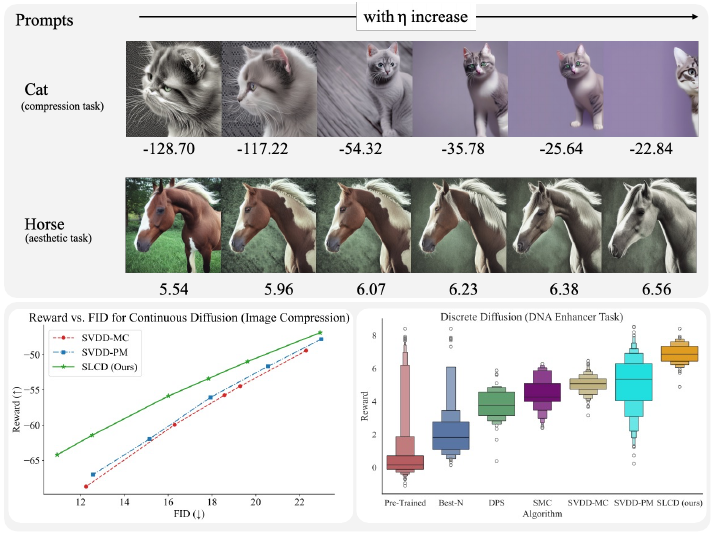}
    \caption{An overview of the main experimental results. \textbf{Top:} Qualitative examples for continuous diffusion image tasks (image compression and aesthetic maximization). Relaxing the KL constraint at test time (larger $\eta$) consistently increases the score. \textbf{Bottom left:} \ALG{} stays closer to the initial image distribution (lower FID score) for the same reward. \textbf{Bottom right:} \ALG{} is likewise effective at controlling discrete diffusion models.}
    \label{fig:experimental-results}
\end{figure}

In this work, we ask the following question: \emph{can we fine-tune diffusion models to optimize a given objective function with simple computation oracles such as regression and classification}, i.e., the standard supervised learning style computation oracles. We provide an affirmative answer to this question (\cref{fig:experimental-results}). We view fine-tuning diffusion model as a controllable generation problem where we train a guidance model -- typically a lightweight small classifier, to guide the pre-trained diffusion model during the inference time. Specifically, we frame the optimization problem as a KL-regularized reward maximization problem where our goal is to optimize the diffusion model to maximize the given reward while staying close to the pre-trained model. Prior work such as SVDD \citep{uehara2024feedback} also studied a similar setting where they also train a guidance model to guide the pre-trained diffusion model in inference time. However SVDD's solution is sub-optimal, i.e., it does not guarantee to find the optimal solution of the KL-regularized reward maximization objective. We propose a new approach \ALG{}: Supervised Learning based Controllable Diffusion, which \textbf{iteratively} refines a classifier using the feedback from the reward functions, on the \textbf{online data} generated by the classifier itself guiding the pre-trained diffusion model. The key computation primitive of \ALG{} is supervised learning, i.e., each iteration it performs multi-class classification on the current collected data. It also collects online data via the standard classifier-guided denoising process. \ALG{} is entirely motivated from the concept of classifier-guidance but with a novel idea on how to learn the \textbf{optimal classifier}, where the optimality means that generated the distribution from the classifier guiding the base model can provably converge to the optimal target distribution. 

Theoretically, for continuous diffusion, we demonstrate that via a reduction to no-regret learning \citep{shalev2012online}, \ALG{} finds a near optimal solution to the KL-regularized reward maximization objective. In other words, \ALG{} manages to balance the reward maximization and KL minimization. Our analysis is motivated from the classic imitation learning (IL) algorithms DAgger \citep{ross2011reduction} and AggreVaTe(d) \citep{ross2014reinforcement,sun2017deeply} which frame IL as an iterative classification procedure with the main computation primitive being classification. Our theory shows that as long as we can achieve no-regret on the sequence of classification losses constructed during the training, the learned classifier can guide the pre-trained diffusion model to generate a near optimal distribution. 

Experimentally, on two types of applications to image generation with continuous diffusion and biological sequence generation with discrete diffusion, we find \ALG{} consistently outperforms other baselines on reward and inference speed, while maintaining a lower divergence from the base model.  
Overall, \ALG{} serves a simple solution to the problem of fine-tuning both continuous and discrete diffusion models to optimize a given KL-regularized reward function. %

%% file: sec/related.tex
\section{Related Work}

There has been significant interest in controllable generation of diffusion models, starting from \cite{dhariwal2021diffusion} which introduced classifier guidance, to then \cite{ho2022classifierfree} which introduced classifier-free guidance. These methods paved the way for further interest in controllable generation, in particular when there is an objective function to optimize. First demonstrated by \cite{black2024ddpo,fan2023DPOK}, RL fine-tuning of diffusion models has grown in popularity with works such as \cite{clark2024directly,prabhudesai2023aligning} which use direct backpropagation to optimize the reward function. However, these methods can lead to mode collapse of the generations and overfitting. Further works then focused on maximizing the KL-constrained optimization problem which regularizes the generation process to the base model \citep{uehara2024feedback} but suffer either from needing special memoryless noise schedulers \citep{domingoenrich2025adjoint} or needing to control the initial distribution of the diffusion process \citep{uehara2024finetuning} to avoid a bias problem. Our approach does not need to do any of these modification. \cite{li2024derivative} proposed a method to augment the decoding process and avoid training the underlying base model, but yield an increase in compute time. Their practical approach also does not guarantee to learn the optimal distribution.  
More broadly, \cite{mudgal2023controlled,zhou2025q} investigate token-level classifier guidance for KL-regularized controllable generation of large language models. \cite{zhou2025q} also demonstrated that to optimize a KL-regularized RL objective in the context of LLM text generation, one just needs to perform no-regret learning. While our approach is motivated by the Q\# approach from \cite{zhou2025q},  
we tackle score-based guidance for diffusion models in a continuous latent space and continue time, where the space of actions form an infinite set--making algorithm design and analysis %
substantially more difficult than the setting of discrete token and discrete-time in prior LLM work.

%% file: sec/preliminaries.tex
\section{Preliminaries\label{sec:preliminaries}}

\subsection{Diffusion models}
Given a data distribution $q_0$ in $\RR^d$, the forward process of a diffusion model (\cite{song2021scorebased}) adds noise to a sample $\bar\xb_0 \sim q_0$ iteratively, which can be modeled as the solution to a stochastic differential equation (SDE):
\begin{equation}
  \label{eq:forward-sde}
  \d \bar\xb = \hb(\bar\xb,\tau) \d \tau + g(\tau) \d \bar\wb, \quad \bar\xb_0 \sim q_0, \quad \tau \in [0,T]
\end{equation}
where $\{\bar\wb_\tau\}_\tau$ is the standard Wiener process, $\hb(\cdot, \cdot): \RR^d \times [0,T] \to \RR^d$ is the drift coefficient and $g(\cdot): [0,T] \to \RR$ is the diffusion coefficient. We use $q_\tau(\cdot)$ to denote the probability density function of $\bar\xb_\tau$ generated according to the forward SDE in equation~\eqref{eq:forward-sde}. We assume $\fb$ and $g$ satisfy certain conditions s.t. $q_T$ converges to $\Ncal(0,I)$ as $T \to \infty$. For example, if \cref{eq:forward-sde} is chosen to be Ornstein–Uhlenbeck process, $q_T$ converges to $\Ncal(0,1)$ exponentially fast.

The forward process \eqref{eq:forward-sde} can be reversed by:
\begin{equation}
  \label{eq:reverse-sde}
  \d \xb = \left[-\hb(\xb,T-t) + g^2(T-t) \nabla \log q_{T-t}(\xb)\right]\d t + g(T-t) \d \wb, \quad \xb_0 \sim q_T, \quad t \in [0,T],
\end{equation}
where $\{\wb_t\}_t$ is the  Wiener process. %
It is known \citep{anderson1982reverse} that the forward process \eqref{eq:forward-sde} and the reverse process \eqref{eq:reverse-sde} have the same marginal distributions. %
To generate a sample, we can sample $x_0 \sim q_T$, and run the above SDE from $t = 0$ to $t = T$ to get $x_T$. In practice, one can start with $\xb_0 \sim \Ncal(0,I)$ (an approximation for $q_T$) and use numerical SDE solver to approximately generate $x_T$, such as the generation processes from DDPM \citep{ho2020denoising} or DDIM \citep{song2021denoising}.

\subsection{Controllable generation}
In certain applications, controllable sample generated from some target conditional distribution is preferable. This can be achieved by adding guidance to the score function. In general, the reverse SDE with guidance $\fb(\cdot,\cdot)$ is:
\begin{equation}
  \label{eq:reverse-guided-sde}
  \d \xb = \left[-\hb(\xb,T-t) + g^2(T-t) \left(\nabla\log q_{T-t}(\xb) + \fb(\xb,t)\right)\right]\d t + g(T-t) \d \wb.
\end{equation}
For convenience, for all $0\le s\le t \le T$, we use
\begin{equation}
  P^{\fb}_{s\to t}(\cdot | p)
\end{equation}
to denote the marginal distribution of $\xb_t$, the solution to~\eqref{eq:reverse-guided-sde} with initial conditional $\xb_s \sim p$. In the remaining of this paper, we may abuse the notation and use $P^{\fb}_{s\to t}(\cdot | \xb')$ to denote a deterministic initial condition $\Pr\left[\xb_s = \xb'\right]=1$.
In particular, we use $P^{\text{prior}}_{s\to t}(\cdot | p)$ to denote the special case that $\fb \equiv 0$.\looseness=-1

\subsection{Reward guided generation}
In this paper, we aim to generate samples that maximize some reward function $r(x)\in [-R_{\max}, 0]$,  while not deviating too much from the base or \textbf{prior} distribution $q_0$. We consider the setting where we have access to the score function of the prior $q_0$ (e.g., $q_0$ can be modeled by a pre-trained large diffusion model). \looseness=-1

Formally, our goal is to find a distribution $p$ that solves the following optimization problem:
\begin{equation}
  \label{eq:kl-regularized-obj}
  \max_{p}\EE_{\xb\sim p}[r(\xb)] - \frac{1}{\eta}\text{KL}(p\|q_0)
\end{equation}
for some $\eta > 0$ which controls the balance between optimizing reward and staying close to the prior.
It is known \citep{ziebart2008maximum} that $p^{\star}$, the optimal solution to the optimization in~\eqref{eq:kl-regularized-obj}, satisfies
\begin{equation}
  \label{eq:target-dist}
  p^{\star}(\xb) = \frac{1}{Z}q_0(\xb) \exp(\eta r(\xb)),
\end{equation}
where $Z>0$ is the normalization factor.
Prior work treats this as a KL-regularized RL optimization problem. However as we mentioned, to ensure the optimal solution of the KL-regularized RL formulation to be equal to $p^\star$, one need to additionally optimize the state distribution $q_T$ (e.g., via another diffusion process), or need to design special memory less noise scheduling \citep{domingoenrich2025adjoint}. We aim to show that we can learn $p^\star$ in a provably optimal manner \emph{without relating the setting to RL or stochastic control}, thus eliminating the needs of optimizing the initial distribution, modifying the noise schedulers, or using complicated RL solvers such as PPO \citep{schulman2017proximal}. \looseness=-1

%% file: sec/algorithm.tex
\section{Algorithm}
\label{sec:algorithm}

\begin{figure}[t]
    \centering
    \includegraphics[width=\linewidth]{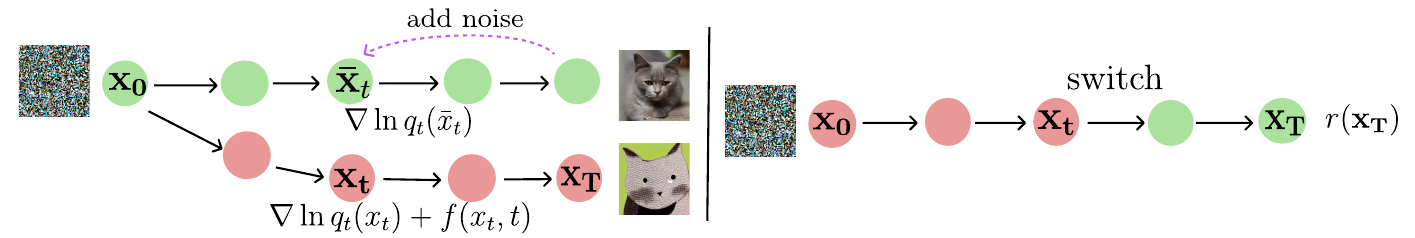}
    \caption{Covariate shift (left) and data collection in our approach (right). The left figure illustrates covariate shift. In the offline naive approach, classifier will be trained under the green samples. However in inference time, the classifier will be applied at the red samples -- samples generated by using the classifier itself as guidance. The difference in green samples (training distribution) and red samples (testing distribution) is the covariate shift. 
    Our approach (right) mitigates this by iteratively augmenting the training set with samples drawn from guided diffusion. We rollin with the classifier-guided diffusion to get to $\xb_t$. We then 
    rollout with the prior's score function to get to $x_T$ and query a reward $r(\xb_T)$. The triple $(t,\xb_t, r(\xb_T))$ will be used to refine the classifier.    }
    \label{fig:dagger}
\end{figure}

We introduce a binary label $y\in \{0,1\}$ and denote the classifier $p(y = 1 | \xb) := \exp( \eta r(\xb))$ (recall that we assume reward is negative). The introduction of the binary label and the classifier allows us to rewrite the target distribution as the posterior distribution given $y=1$:
\begin{align*}
    p(\xb | y = 1) \propto q_0(\xb) p(y = 1 | \xb) = q_0(\xb)\exp( \eta r(\xb)).
\end{align*}

Given this formulation, the naive approach is to model $p(\xb | y = 1)$ via the standard classifier-guided diffusion process. In other words, we generate a labeled dataset $\{(\xb, y)\}$  where $\xb$ is from the prior $\xb\sim q_0$, and the label is sampled from the Bernoulli distribution with mean $\exp(\eta r(\xb))$, i.e., $y \sim p( y | \xb)$. Once we have this data, we can add noise to $\xb$, train a time-dependent classifier that predicts the noised sample to its label $y$. Once we train the classifier, we use its gradient to guide the generation process as shown in \cref{eq:reverse-guided-sde}. 

While this naive approach is simple, this approach can fail due to the issue of \emph{covariate shift} -- the training distribution (i.e., $q_t$ -- the distribution of $\bar \xb_t$) used for training the classifier is different from the test distribution where the classifier is evaluated during generation (i.e. the distribution of samples $x$ generated during the classifier-guided denoising process). In other words, while the classifier can be an accurate classifier under the training distribution, there is no guarantee that it is an accurate classifier under the test distribution -- the distribution that ultimately matters when we perform classifier-guided generation.  This is demonstrated in left figure in \cref{fig:dagger}. In the worst case, the density ratio of the test distribution over the training distribution can be exponential $\exp( R_{\max}  \eta)$, which can be too large to ignore when $\eta$ is large (i.e., KL regularization is weak).\looseness=-1

We propose an iterative approach motivated by DAgger (data aggregation, \cite{ross2011reduction}) to close the gap between the training distribution and test distribution. First with the binary label $y$ and our definition of $p(y=1 | \xb_T) = \exp(\eta r(\xb_T))$ (note $\xb_T$ represents the generated image), we can show that the classifier $p(y = 1 | \xb_t)$ for any $t\in [0,T)$ takes the following form: 
\begin{align}
\label{eq:classifier}
    p( y = 1 | \xb_t) = \mathbb{E}_{x_T \sim P^{\text{prior}}_{t\to T}(\cdot | \xb_t)} \exp( \eta r( \xb_T ) ).
\end{align} Intuitively $p(y=1 | \xb_t)$ models the expected probability of observing label $y=1$ if we generate $\xb_T$ starting from $\xb_t$ using the reverse process of the pre-trained diffusion model. We defer the formal derivation to Appendix~\ref{app:derive-classifier} which relies on proving that the forward process and backward process of a diffusion model induce the same conditional distributions.

We take advantage of this closed-form solution of the classifier, and propose to model the classifier via a \emph{distributional approach} \citep{zhou2025q} .  Particularly, define $r\sim R^{\text{prior}}(\cdot | \xb_t, t)$ as the distribution of the reward of a $\xb_T \sim P^{\text{prior}}_{t\to T}(\cdot | \xb_t)$. The classifier $p( y = 1 | \xb_t)$ can be rewritten using the reward distribution $R^{\text{prior}}(\cdot | \xb_t, t)$: \looseness=-1
\begin{align}
    p( y = 1 | \xb_t) := \mathbb{E}_{r\sim R^{\text{prior}}(\cdot | \xb_t,t)} \exp( \eta \cdot r ).
\label{eq:reward_distribution_classifier}
\end{align}
Our goal is to learn a reward distribution $\hat R$ to approximate $R^{\text{prior}}$ and use $\hat R$ to approximate the classifier as $p(y=1 | \xb_t) \approx \mathbb{E}_{r\sim \hat R(\cdot | \xb_t, t)} \exp( \eta r)$. This distributional approach allows us to take advantage of the closed form of the classifier in \cref{eq:classifier} (e.g., there is no need to learn the exponential function $\exp( \eta r)$ in the classifier). \cref{alg:main-alg} describes an iterative learning approach for training such a distribution $\hat R(\cdot | \xb_t, t)$ via supervised learning (e.g., maximum likelihood estimation (MLE)). 

\input{algorithms/algo-pseudo}

Inside iteration $n$, given the current reward distribution $\hat R^n$, we define the score of the classifier $\fb^n$ as:\looseness=-1
\begin{align}
\label{eq:fn}
    \forall t, x_t: \; \fb^n(\xb_t, t) := \nabla_{\xb_t} \ln \left( \mathbb{E}_{r\sim \hat R^n(\cdot | \xb_t, t)}\exp( \eta \cdot r  )\right).
\end{align} 
We use $\fb^n$ to guide the prior to generate an additional training dataset $D^n:=\{(t, \xb_t, r)\}$ of size $M$, where 
\begin{align}
\label{eq:data-collection}
\underbrace{t\sim \text{Uniform}(T), \xb_0 \sim \Ncal(0,I),  \xb_t \sim P^{\fb^n}_{0\to t}(\cdot | \xb_0)}_{\text{\textbf{Roll in} with the  score of the latest classifier $\fb^n$ as guidance}}, \;\; \underbrace{\xb_T \sim P^{\text{prior}}_{t\to T}( \cdot | \xb_t ), \text{and } r = r(\xb_T)}_{\text{\textbf{Roll out} with the prior to collect reward}}.
\end{align}
Note that the roll-in process above \text{simulates the inference procedure} -- $\xb_t$ is an intermediate sample we would generate if we had used $\fb^n$ to  guide the prior in inference. %
The rollout procedure collects reward signals for $\xb_t$ which in turn will be used for refining the reward distribution estimator $\hat R(\cdot | \xb_t)$. This procedure is illustrated in the right figure in \cref{fig:dagger}. We then aggregate $D^n$ with all the prior data and re-train the distribution estimator $\hat R$ using the aggregate data via supervised learning, i.e., maximum likelihood estimation:
\begin{align}
\label{eq:R-n+1}
    \hat R^{n+1} \in \argmax_{R \in \Rcal} \sum_{i=1}^n \sum_{(t,\xb_t,r)\in D^i}  \ln R( r | \xb_t, t),
\end{align}
where $\Rcal$ is the class of distributions. This rollin-rollout procedure is illustrated in \cref{fig:dagger}.
We iterate the above procedure until we reach a point where $\hat R^n(\cdot | \xb_t, t)$ is an accurate estimator of the true model $R^{\text{prior}}(\cdot | \xb_t, t)$ under distribution induced by the generation process of guiding the prior using $\fb^n$ itself. Similar to DAgger's analysis, we will show in our analysis section that a simple no-regret argument implies that we can reach to such a stage where there is no gap between training and testing distribution anymore. 

\textbf{In the test time}, once we have the score $\fb^{\hat n}$, we can use it to guide the prior to generate samples via the SDE in \cref{eq:reverse-guided-sde}. In practice, sampling procedure from DDPM can be used as the numerical solver for the SDE \cref{eq:reverse-guided-sde}. Another practical benefit of our approach is that the definition of the distribution $R^{\text{prior}}$ and the learned distribution $\hat R$ are independent of the guidance strength parameter $\eta$. This means that in practice, once we learned the distribution $\hat R$, we can adjust $\eta$ during inference time as shown in \cref{eq:fn} without re-training $\hat R$.

\begin{remark} [Modeling the one-dimensional distribution as a classifier]
We emphasize that from a computation perspective, our approach relies on a simple supervised learning oracle to estimate the \textbf{one dimensional conditional distribution $R(r | \xb_t, t)$}. In our implementation, we use histogram to model this one-dimensional distribution, i.e., we discretize the range of reward $[-R_{\max}, 0]$ into finite number of bins, and use standard \textbf{multi-class classification oracle} to learn $\hat R$ that maps from $\xb_t$ to a distribution over the finite number of labels. Thus, unlike prior work that casts controllable diffusion generation as an RL or stochastic control problem, our approach eliminates the need to talk about or implement RL, and instead entirely relies on standard classification and can be seamlessly integrated with any existing implementation of classifier-guided diffusion. \looseness=-1
\end{remark}

\begin{remark}[Comparison to SVDD \citep{uehara2024feedback}] The most related work is SVDD. There are two notifiable differences. First SVDD estimates a sub-optimal classifier, i.e., $\mathbb{E}_{r\sim R^{\text{prior}}(\cdot | \xb_t,t)}[ r ]$. The posterior distribution in their case is proportional to $q_0(\xb_T)\cdot \mathbb{E}_{r\sim R^{\text{prior}}(\cdot | \xb_T,T)}[ r ]$ which is clearly not equal to the  target distribution. Second, SVDD does not address the issue of distribution shift and it trains the classifier only via offline data collected from the prior alone. 
\end{remark}

We note that our method can also be adapted to discrete diffusion tasks as seen in \cref{sec:experiments}. We refer the reader to \cref{sec:training} for more details.

%% file: algorithms/algo-pseudo.tex
\begin{algorithm}[t]
\caption{Controllable diffusion via iterative supervised learning (\ALG{})\label{alg:main-alg}}
\begin{algorithmic}
\State Initialize $\hat R^1$.
\For{$n = 1, \ldots, N$}
\State set $\fb^n$ as~\cref{eq:fn}.
\State collect an additional training dataset $D^n$ following~\cref{eq:data-collection}.
\State train $\hat R^{n+1}$ on $\bigcup_{i=1}^n D^i$ according to~\cref{eq:R-n+1}.
\EndFor
\State \textbf{Return} $\fb^{\hat n}$, the best of $\{\fb^{1},\ldots,\fb^{N}\}$ on validation.     
\end{algorithmic}
\end{algorithm}

%% file: sec/analysis.tex
\section{Analysis}
\label{sec:analysis}
In this section, we provide performance guarantee for the sampler returned by \cref{alg:main-alg}. We use KL divergence of the generated data distribution $P^{\fb^{\hat n}}_{0\to T}(\cdot | \Ncal(0,I))$ from the target distribution $p^{\star}$ to measure its quality. At a high level, the error comes from two sources:
\begin{itemize}[leftmargin=*]
    \item \textbf{starting distribution mismatch:} in the sampling process, we initialize the SDE \cref{eq:reverse-guided-sde} with samples from $\Ncal(0,I)$, not the ground-truth $q_T(\cdot | y=1)$. However, under proper condition, $q_T(\cdot | y=1)$ converges to $\Ncal(0,I)$ as $T\to\infty$ (see Lemma 3 of \cite{chen2025convergence}). In particular, when \cref{eq:reverse-guided-sde} is chosen to be the OU process, $q_T(\cdot | y=1)$ converges at an exponential speed: $\text{KL}(\Ncal(0,I)\| q_T(\cdot | y=1)) = O(e^{-2T})$. 
    \item \textbf{estimation error of the guidance:} the estimated guidance $\fb^{\hat n}$ is different from the ground truth $\nabla_{\xb_t}\ln p(y=1|\xb_t)$. But the error is controlled by the regret of the no-regret online learning. 
\end{itemize}
We assume realizability:
\begin{assum}[realizability]\label{lem:realizability}
  $R^{\text{prior}} \in \Rcal$
\end{assum}

Our analysis relies on a reduction to no-regret online learning. Particularly, we assume we have no-regret property on the following log-loss. Note $M$ as the side of the each online dataset $D^i$.
\begin{assum}[No-regret learning]\label{assum:no-regret} The sequence of reward distribution $\{\hat R^i\}$ satisfies the following inequality:
\begin{align*}
    \frac{1}{NM}\sum_{i=1}^N \sum_{(t,\xb_t,r)\in D^i}\ln\frac{1}{\hat R^i(r|\xb_t,t)}
    -
    \min_R \frac{1}{NM}\sum_{i=1}^N \sum_{(t,\xb_t,r)\in D^i}\ln\frac{1}{R(r|\xb_t,t)}
  \le
  \gamma_N.
\end{align*} where the average regret $\gamma_N = o(N)/N$ shrinks to zero when $N\to\infty$.
\end{assum}
No-regret online learning for the $\log$ is standard in the literature \citep{cesa2006prediction,foster2021statistical,wang2024central,zhou2025q}. Our algorithm implements the specific no-regret algorithm called Follow-the-regularized-leader (FTRL) \citep{shalev2012online,suggala2020online} where we optimize for $\hat R^i$ on the aggregated dataset. Follow-the-Leader type of approach with random perturbation can even achieve no-regret property for non-convex optimization \citep{suggala2020online}. This data aggregation step and the reduction to no-regret online learning closely follows DAgger's analysis \citep{ross2011reduction}.

Under certain condition, the marginal distribution $q_T$ defined by the forward SDE \cref{eq:forward-sde} converges to some Gaussian distribution rapidly (see Lemma 3 of \cite{chen2025convergence}). For simplicity, we make the following assumption on the convergence:
\begin{assum}[convergence of the forward process]\label{assum:convergence-forward}
$\text{KL}(\Ncal(0,I)\| q_T(\cdot | y=1)) \le \epsilon_T$.
\end{assum} For OU processes, $\epsilon_T$ shrinks in the rate of $\exp(-T)$. 
We assume the reward distribution class satisfies certain regularity conditions, s.t. the estimation error of the classifier controls the score difference: 
\begin{assum}\label{assum:reward-class-condition}
  There exists $L>0$, s.t. for all $R, R' \in \Rcal$, and $\xb,t$:
  \begin{align*}
    &\left\|\nabla_{\xb} \ln \left( \mathbb{E}_{r\sim R(\cdot | \xb, t)}\exp( \eta \cdot r  )\right) - \nabla_{\xb} \ln \left( \mathbb{E}_{r\sim R'(\cdot | \xb, t)}\exp( \eta \cdot r  )\right)\right\|_2 \\
    \le & 
    L\left|\mathbb{E}_{r\sim R(\cdot | \xb, t)}\exp( \eta \cdot r  )- \mathbb{E}_{r\sim R'(\cdot | \xb, t)}\exp( \eta \cdot r  )\right|.
  \end{align*}
\end{assum}
Standard diffusion models with classifier guidance train a time-dependent classifier and use the score function of the classifier to control image generation \citep{song2021scorebased,dhariwal2021diffusion}. Such assumption is crucial to guarantee the quality of class-conditional sample generation. We defer a more detailed discussion on Assumption~\ref{assum:reward-class-condition} to Appendix~\ref{sec:classifier-score-assumption}. In general such an assumption holds when the functions satisfy certain smoothness conditions.

\begin{theorem}
  Suppose Assumption~\ref{lem:realizability},~\ref{assum:no-regret},~\ref{assum:convergence-forward}, and~\ref{assum:reward-class-condition} hold.
  There exists $\hat n\in\{1,\ldots, N\}$, s.t. $\fb^{\hat n}$ specified by \cref{alg:main-alg} satisfies:
  \begin{align*}
    \EE\left[\KL\left(P^{\fb^{\hat n}}_{0\to T}(\cdot | \Ncal(0,I))\| p^{\star}\right)\right]
    \le 
    \epsilon_T + \frac{1}{2}T\|g\|_{\infty}^2L^2\gamma_N.
  \end{align*} where the expectation is with respect to the randomness in the whole training process, and $g$ is the diffusion coefficient defined in \cref{eq:forward-sde}.
\end{theorem}%
Since $P^{\fb^{\hat n}}_{0\to T}(\cdot | \Ncal(0,I))$ models the distribution of the generated samples when using $\fb^{\hat n}$ to guide the prior, the above theorem proves that our sampling distribution is close to the target $p^\star$ under KL. Note that $\epsilon_T$ decays in the rate of $\exp(-T)$ when the forward SDE is an OU process, and $\gamma_N$ decayes in the rate of $1/\sqrt{N}$ for a typical no-regret algorithm such as Follow-the-Learder \citep{shalev2012online,suggala2020online}.

%% file: sec/experiments.tex
\section{Experiments}\label{sec:experiments}

We compare \ALG{} to a variety of training‐free and value‐guided sampling strategies across four tasks. For Best-of-N, we draw $N$ independent samples from the base diffusion model and keep the one with the highest reward. Diffusion Posterior Sampling (DPS) is a classifier-guidance variant originally for continuous diffusion \citep{chung2023diffusion}, here adapted to discrete diffusion via the state-of-the-art method of \citet{nisonoff2025unlocking}.  Sequential Monte Carlo (SMC) methods \citep{del2014particle,wu2023practical,trippe2022diffusion} use importance sampling across whole batches to select the best sample.  SVDD-MC \citep{li2024derivative} instead evaluates the expected reward of $N$ candidates under an estimated value function, while SVDD-PM uses the true reward for each candidate with slight algorithm modifications. We evaluate on (i) image compression (negative file size) and (ii) image aesthetics (LAION aesthetic score) using Stable Diffusion v1.5 \citep{rombach2022stable}, as well as on (iii) 5' untranslated regions optimized for mean ribosome load \citep{sample2019human,sahoo2024simple} and (iv) DNA enhancer sequences optimized for predicted expression in HepG2 cells via the Enformer model \citep{avsec2021effective}.

\begin{wrapfigure}[16]{r}{0.5\textwidth}  
  \vspace{-20pt}
  \centering
  \includegraphics[width=\linewidth]{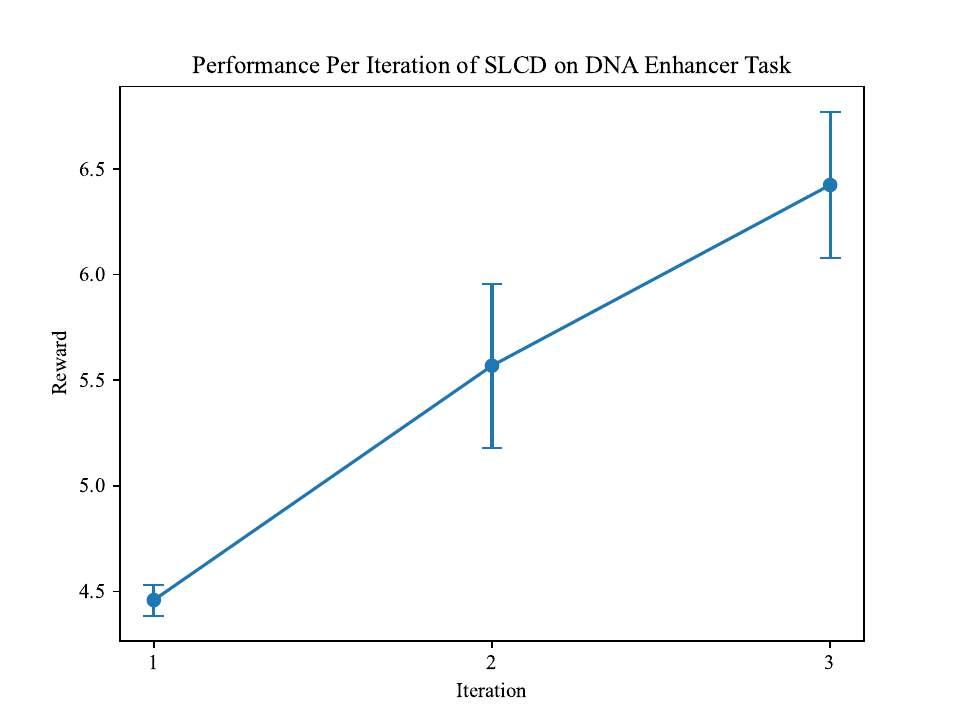}
    \vspace{-0.2in}
  \caption{Reward vs.\ number of iterations of \ALG{}. The reward increases as the restart‐state distribution becomes richer, and only a few iterations are needed to reach good performance.}
  \label{fig:reward_vs_iter}
\end{wrapfigure}

In line with \cite{li2024derivative}, we compare the top 10 and 50 quantiles of a batch of generations, in \cref{tab:main_results}. We compare to these methods as, like \ALG{}, all of these methods do not require training of the base model. \looseness=-1

Overall, we see that \ALG{} consistently outperforms the baseline methods while requiring nearly the same inference time as the base model, and omitting the need for multiple MC samples during each diffusion step.

These four tasks jointly cover two primary application domains of diffusion models: image generation and biological sequence generation, providing a comprehensive assessment of controllable diffusion methods.

\subsection{Reward Comparison}

We compare the reward of \ALG{} to the baseline methods in \cref{tab:main_results}. We see that \ALG{} is able to consistently achieve higher reward than SVDD-MC and SVDD-PM, and the other baseline methods in all four tasks. The margin of improvement is most pronounced in settings where the classifier closely approximates the true reward,most notably the image compression task, where \ALG{} nearly attains the optimal reward. To further see the performance of \ALG{}, we plot the reward distribution of \ALG{} and the baseline methods in \cref{fig:boxplot_rna} and \cref{fig:experimental-results}. We observe that \ALG{} produces a more tightly concentrated reward distribution with a higher median reward than the baseline methods, while still maintaining generation diversity, as shown in \cref{fig:experimental-results} with a lower FID score than baseline methods.

\subsection{Qualitative results}

\begin{figure}[t]
  \centering
  \includegraphics[width=\textwidth]{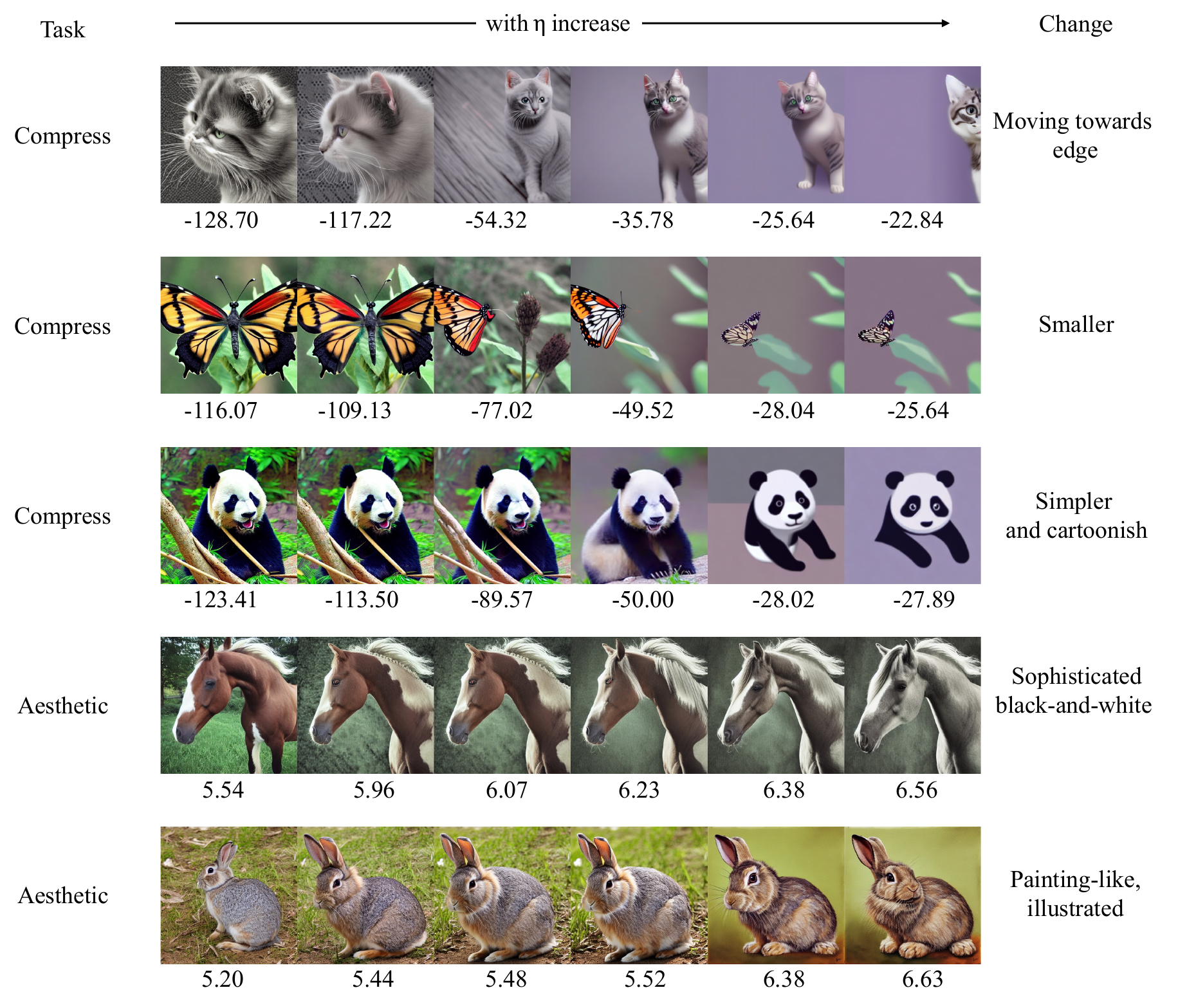}
  \vspace{-0.2in}
  \caption{Images generated by \ALG{} with varying $\eta$ values and their rewards . The first column shows results from the base model, SD1.5, which corresponds to our method with $\eta = 0$. As $\eta$ increases, the KL penalty is relaxed, allowing the generated images to be more strongly optimized for the reward function, and consequently, they diverge further from the base model’s original distribution.}
  \label{fig:images}
\end{figure}

We present generated images from \ALG{} in \cref{fig:images}. For the compression task, we observe three recurring patterns: some images shift the subject toward the edges of the frame, others reduce the subject’s size, and some simplify the overall scene to reduce file size. For the aesthetic task, the outputs tend to take on a more illustrated appearance, often reflecting a variety of artistic styles.

As $\eta$ increases, the KL constraint is relaxed, enabling a controlled trade-off between optimizing for the reward function and staying close to the base model’s distribution. Notably, even under strong reward guidance (i.e., with larger $\eta$), our method consistently maintains a high level of diversity in the generated outputs.

\begin{table}[!t]
  \centering
    \caption{Top 10 and 50 quantiles of the generated samples for each algorithm (with 95\% confidence intervals). Higher is better. \ALG{} consistently outperforms the baseline methods.
}
  \resizebox{\textwidth}{!}{
\Large
\begin{tabular}{c|c|ccccccg} \toprule
Domain  & Quantile & Pre-Train & Best-N & DPS  & SMC & SVDD-MC & SVDD-PM & \textbf{\ALG{}} \\ \midrule
Image: Compress 
& 50\% & 
-101.4\textsubscript{\scriptsize{$\pm$ 0.22}} & 
-71.2\textsubscript{\scriptsize{$\pm$ 0.46}}  &   
-60.1\textsubscript{\scriptsize{$\pm$ 0.44}}  &  
-59.7\textsubscript{\scriptsize{$\pm$ 0.4}} & 
-54.3\textsubscript{\scriptsize{$\pm$ 0.33}} & 
-51.1\textsubscript{\scriptsize{$\pm$ 0.38}} & 
\textbf{-13.60\textsubscript{\scriptsize{$\pm$ 0.79}}} \\

& 10\% & 
-78.6\textsubscript{\scriptsize{$\pm$ 0.13}} & 
-57.3\textsubscript{\scriptsize{$\pm$ 0.28}} &  
-61.2\textsubscript{\scriptsize{$\pm$ 0.28}} &   
-49.9\textsubscript{\scriptsize{$\pm$ 0.24}}  &   
-40.4\textsubscript{\scriptsize{$\pm$ 0.2}} & 
-38.8\textsubscript{\scriptsize{$\pm$ 0.23}} & 
\textbf{-11.05\textsubscript{\scriptsize{$\pm$ 0.41}}} \\ \midrule

Image: Aesthetic 
& 50\% & 
5.62\textsubscript{\scriptsize{$\pm$ 0.003}} & 
6.11\textsubscript{\scriptsize{$\pm$ 0.007}} & 
5.61\textsubscript{\scriptsize{$\pm$ 0.009}} & 
6.02\textsubscript{\scriptsize{$\pm$ 0.004}} & 
5.70\textsubscript{\scriptsize{$\pm$ 0.008}} & 
6.14\textsubscript{\scriptsize{$\pm$ 0.007}} & 
\textbf{6.31\textsubscript{\scriptsize{$\pm$ 0.061}}} \\

& 10\% &  
5.98\textsubscript{\scriptsize{$\pm$ 0.002}} &  
6.34\textsubscript{\scriptsize{$\pm$ 0.004}} &  
6.00\textsubscript{\scriptsize{$\pm$ 0.005}} & 
6.28\textsubscript{\scriptsize{$\pm$ 0.003}} & 
6.05\textsubscript{\scriptsize{$\pm$ 0.005}} & 
6.47\textsubscript{\scriptsize{$\pm$ 0.004}} & 
\textbf{6.59\textsubscript{\scriptsize{$\pm$ 0.077}}} \\ \midrule

Enhancers (DNA) 
& 50\% & 
0.121\textsubscript{\scriptsize{$\pm$ 0.033}} & 
1.807\textsubscript{\scriptsize{$\pm$ 0.214}} & 
3.782\textsubscript{\scriptsize{$\pm$ 0.299}} &  
4.28\textsubscript{\scriptsize{$\pm$ 0.02}} & 
5.074\textsubscript{\scriptsize{$\pm$ 0.096}} & 
5.353\textsubscript{\scriptsize{$\pm$ 0.231}}  & 
\textbf{7.403\textsubscript{\scriptsize{$\pm$ 0.125}}}  \\

& 10\% & 
1.396\textsubscript{\scriptsize{$\pm$ 0.020}} & 
3.449\textsubscript{\scriptsize{$\pm$ 0.128}} & 
4.879\textsubscript{\scriptsize{$\pm$ 0.179}} & 
5.95\textsubscript{\scriptsize{$\pm$ 0.01}} & 
5.639\textsubscript{\scriptsize{$\pm$ 0.057}} & 
6.980\textsubscript{\scriptsize{$\pm$ 0.138}} & 
\textbf{7.885\textsubscript{\scriptsize{$\pm$ 0.231}}} \\ \midrule

5'UTR (RNA) 
& 50\% & 
0.406\textsubscript{\scriptsize{$\pm$ 0.028}} & 
0.912\textsubscript{\scriptsize{$\pm$ 0.023}} & 
0.426\textsubscript{\scriptsize{$\pm$ 0.073}} & 
0.76\textsubscript{\scriptsize{$\pm$ 0.02}} & 
1.042\textsubscript{\scriptsize{$\pm$ 0.008}} & 
1.214\textsubscript{\scriptsize{$\pm$ 0.016}} & 
\textbf{1.313\textsubscript{\scriptsize{$\pm$ 0.024}}} \\

& 10\% & 
0.869\textsubscript{\scriptsize{$\pm$ 0.017}} & 
1.064\textsubscript{\scriptsize{$\pm$ 0.014}} & 
0.981\textsubscript{\scriptsize{$\pm$ 0.044}} & 
0.91\textsubscript{\scriptsize{$\pm$ 0.01}} & 
1.117\textsubscript{\scriptsize{$\pm$ 0.005}} & 
1.383\textsubscript{\scriptsize{$\pm$ 0.010}} & 
\textbf{1.421\textsubscript{\scriptsize{$\pm$ 0.039}}} \\ \bottomrule
\end{tabular}
  }
  \label{tab:main_results}
\end{table}

\subsection{Fréchet Inception Distance  Comparison}
Both \ALG{} and SVDD baselines allow one to control the output sample reward at test time, but via different control variables: \ALG{} modulates the KL-penalty coefficient $\eta$, while SVDD-MC and SVDD-PM vary the number of Monte Carlo rollouts evaluated at each diffusion step. Since these control parameters can affect sample quality in different ways, we report both the Fréchet Inception Distance~\citep{heusel2017gans} (FID) and reward. For the same reward, \textit{higher} FID indicates that the model generates images stray farther from the base models distribution, a sign of reward hacking. We evaluate these methods in \cref{fig:experimental-results}. That is, the points on the curve form a pareto frontier between reward and FID. \ALG{} is able to achieve a better reward-FID trade-off than SVDD-MC and SVDD-PM.

\begin{figure}[t]
  \begin{minipage}[t]{0.48\textwidth}
    \centering
    \includegraphics[width=\textwidth]{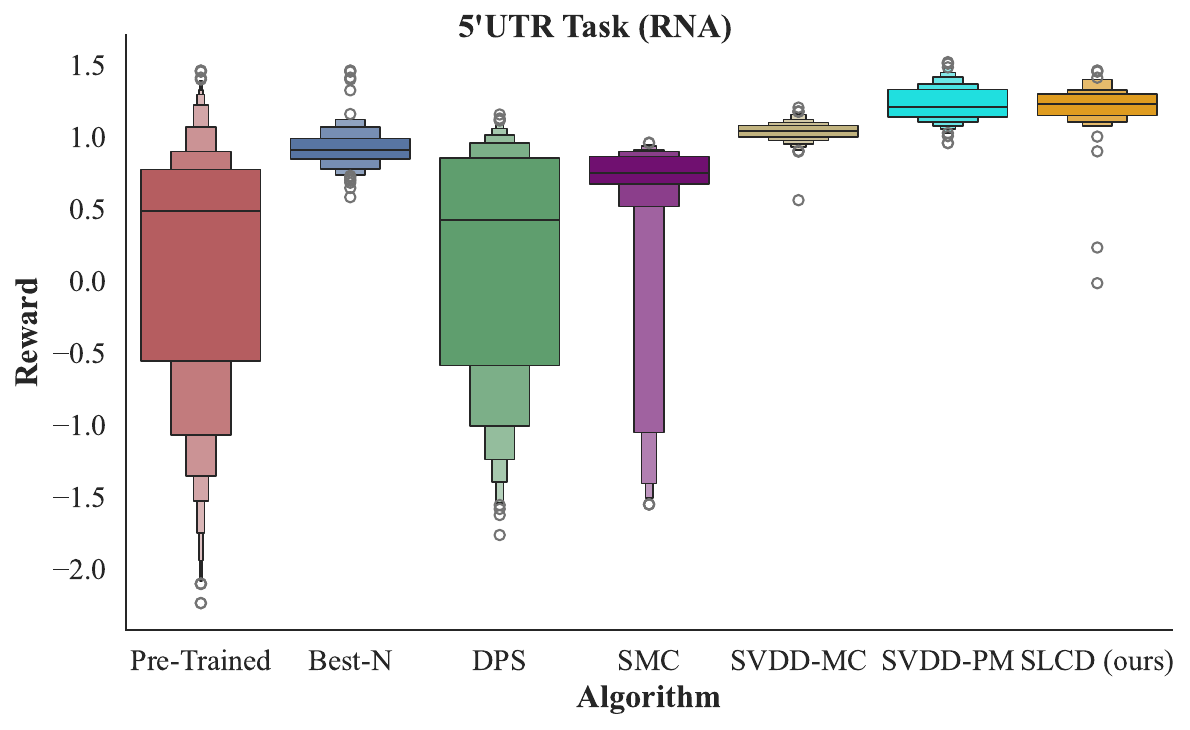}
      \vspace{-0.2in}
    \caption{Distribution of rewards for RNA sequences (5'UTR) across different methods. \ALG{} achieves higher median rewards and better overall distribution compared to baseline approaches.}
    \label{fig:boxplot_rna}
  \end{minipage}
  \hfill
    \begin{minipage}[t]{0.48\textwidth}
  \centering
  \includegraphics[width=\textwidth]{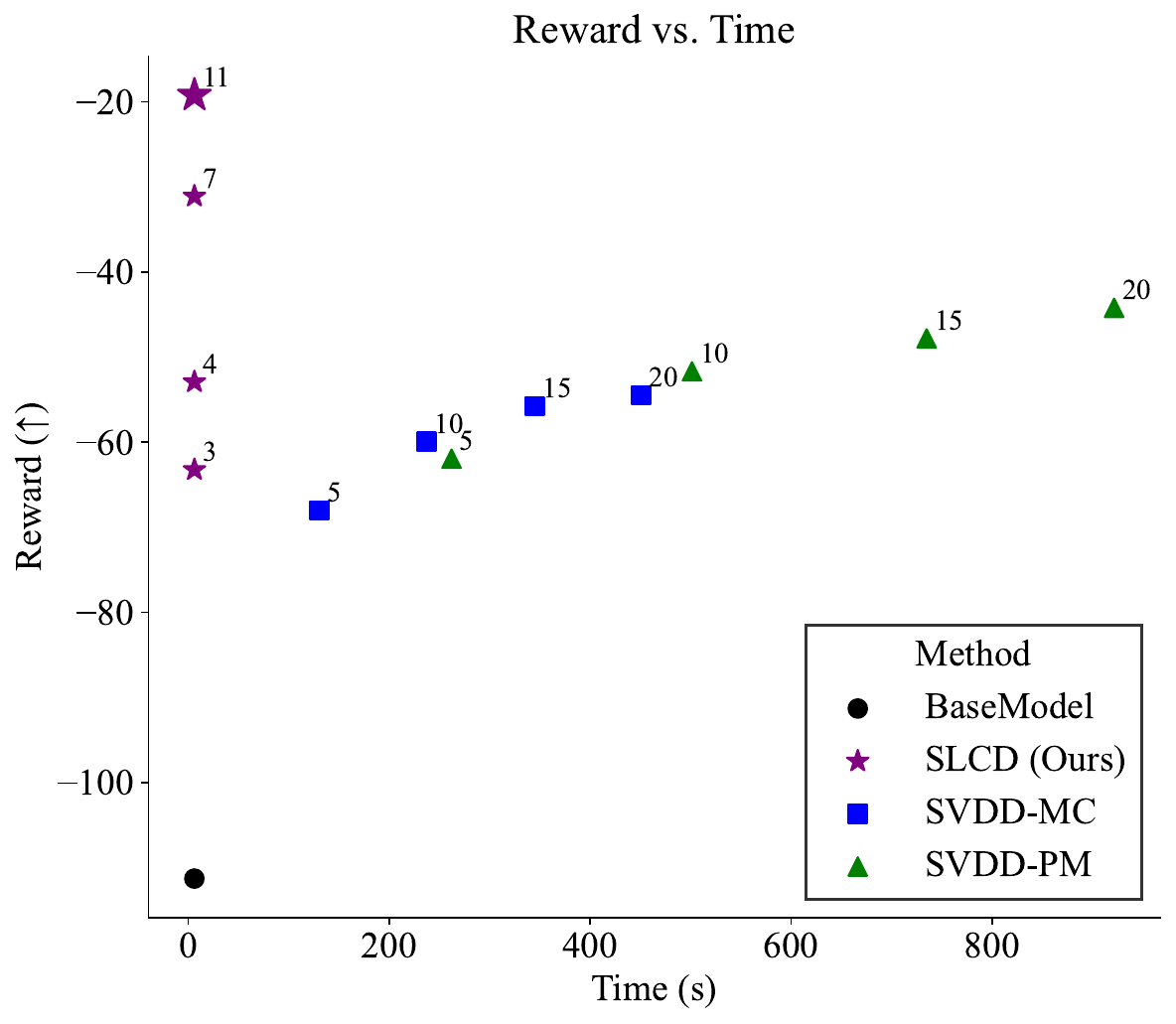}
    \vspace{-0.2in}
  \caption{Reward vs. Inference Time on the Compression Task.
 Numeric labels on the SLCD indicate $\eta$, while those on the SVDD denote the duplication number applied at each step.}
  \label{fig:inference_time}
\end{minipage}
\end{figure}

\subsection{Inference Time Comparison}
An additional advantage of \ALG{} is its negligible inference overhead at test time, even when using higher $\eta$ values to achieve greater rewards. In \cref{fig:inference_time}, we compare the wall-clock generation time per image on an NVIDIA A6000 GPU for \ALG{} against SVDD-MC and SVDD-PM. \ALG{} achieves higher rewards while requiring significantly fewer computational resources and substantially shorter inference times than either baseline. Specifically, \ALG{} takes only 6.06 seconds per image, nearly identical to the base model SD 1.5’s 5.99 seconds.

Importantly, unlike SVDD methods that incur increased computational cost to improve rewards, \ALG{} maintains constant inference time across all $\eta$ values, achieving enhanced performance with no additional computation.

\subsection{Ablation Study}
To elucidate the impact of each training cycle, we vary the number of \ALG{} iterations and plot the resulting reward in \cref{fig:reward_vs_iter}. As additional iterations enrich the state distribution and mitigate covariate shift for the classifier, the reward consistently rises. This confirms that our iterative approach can mitigate covariate shift issue. In practice, we only require a small number of iterations to achieve high reward.

Because the scaling parameter~$\eta$ can be chosen at test time when the distribution $\hat R$ is fully trained, \ALG{} enables test-time control over the KL penalty during inference. By modulating $\eta$ at test-time, practitioners can smoothly trade-off reward against sample quality without retraining the distribution $\hat R$, as demonstrated in \cref{fig:images}.

%% file: sec/conclusion.tex
\section{Conclusion}

In this work, we introduced \ALG{}, a novel and efficient method that recasts the KL-constrained optimization problem as a supervised learning task. We provided theoretical guarantees showing that \ALG{} converges to the optimal KL-constrained solution and how data-aggregation effectively mitigates covariate shift. Empirical evaluations confirm that \ALG{} surpasses existing approaches, all while preserving high fidelity to the base model's outputs and maintaining nearly the same inference time. \looseness=-1

%% file: sec/ack.tex
\begin{ack}
Wen Sun is supported by NSF IIS-2154711,
NSF CAREER 2339395, DARPA LANCER: LeArning Network CybERagents, and the Sloan Research Fellowship.
\end{ack}

\newpage

%% file: sec/appendix.tex
We show the derivation for the classifier in \cref{app:derive-classifier}. We discuss \cref{assum:reward-class-condition} in \cref{sec:classifier-score-assumption}. In \cref{app:proof}, we provide the proof for the main theorem in the main text. Then, we provide additional technical lemmas in \cref{sec:technical-lemmas}. Finally, we provide more details on the training and evaluation of \ALG{} in \cref{sec:training}. We finally provide more samples of the image experiments in \cref{sec:image-samples}.

\section{Derivation for the classifier}
\label{app:derive-classifier}
Our goal is to show \cref{eq:classifier}:
\begin{align*}
  p( y = 1 | \xb_t) = \mathbb{E}_{\xb_T \sim P^{\text{prior}}_{t\to T}(\cdot | \xb_t)} \exp( \eta r( \xb_T ) ).
\end{align*}

Our first step is to derive the classifier in terms of $\{\bar \xb_{\tau}\}_{\tau}$, the forward process defined by \cref{eq:forward-sde}.
By the law of total probability:
\begin{align*}
  p( y = 1 | \bar \xb_{\tau}) = & \frac{p( y = 1, \bar \xb_{\tau})}{q(\bar \xb_{\tau})}
  = \dfrac{\int p( y = 1, \bar \xb_{\tau}, \bar \xb_{0})\d \bar\xb_0}{q(\bar\xb_{\tau})}
  = \dfrac{\int p( y = 1|\bar \xb_{\tau}, \bar \xb_{0})q(\bar \xb_{\tau}, \bar \xb_{0})\d \bar\xb_0}{q(\bar\xb_{\tau})}\\
  = & \int p( y = 1|\bar \xb_{\tau}, \bar \xb_{0})q(\bar \xb_{0} | \bar \xb_{\tau})\d \bar\xb_0.
\end{align*}
By definition, the label $y$ and $\bar \xb_{\tau}$ are independent when conditioning on $\bar \xb_0$, thus $p( y = 1|\bar \xb_{\tau}, \bar \xb_{0}) = p( y = 1|\bar \xb_{0})$ and 
\begin{equation}
\label{eq:classifier-forward}
  p( y = 1 | \bar \xb_{\tau}) 
  =  \EE_{\bar \xb_{0} | \bar \xb_{\tau}}\left[p( y = 1|\bar \xb_{0}) \right]
  = \EE_{\bar \xb_{0} | \bar \xb_{\tau}}\left[\exp(\eta r(\bar \xb_{0})) \right].
\end{equation}

By Lemma~\ref{lem:joint-dist-reverse-forward-sde}, the forward (\cref{eq:forward-sde})and reverse (\cref{eq:reverse-sde}) processes have the same conditional distribution. Thus, in \cref{eq:classifier-forward}, we can substitute $\bar \xb_0$, $\bar \xb_{\tau}$, $q(\bar \xb_{0} | \bar \xb_{\tau})$ with the corresponding components in the reverse process: $\xb_T$, $\xb_{T-\tau}$, $P^{\text{prior}}_{T-\tau\to T}(\xb_T | \xb_{T-\tau})$. Then we complete the proof by setting $\tau = T-t$. 

One crucial property is: \cref{eq:classifier-forward}, the classifier defined in terms of the forward process, only depends on the conditional distribution $\bar \xb_0 | \bar \xb_{\tau}$, not the joint distribution of $(\bar \xb_0, \bar \xb_{\tau})$. This means \cref{eq:classifier}, the classifier defined in terms of the reverse process, is accurate regardless of the marginal distribution of $\xb_t$. 

In fact, when considering the data collection procedure \cref{eq:data-collection}, the roll-in step involves the classifier estimator during training, so the marginal distribution of $\xb_t$ can be very different from $q_{T-t}$, the marginal distribution of the forward process. However, as long as the roll-out step uses the ground truth reverse process, we are using the correct conditional distribution and the target classifier during training is thus unbiased.

\section{Discussion on Assumption~\ref{assum:reward-class-condition}}
\label{sec:classifier-score-assumption}

Assumption~\ref{assum:reward-class-condition} assumes that the estimation for the classifier results in good gradient estimation with a small pointwise error. In this section, we show two sets of sufficient conditions that weaken Assumption~\ref{assum:reward-class-condition} to versions where the estimation errors are evaluated on the marginal distribution of the reverse process. The conditions will depend on:
\begin{itemize}
  \item $\hat R^{\hat n}$: the reward distribution chosen by~\cref{alg:main-alg};
  \item $R^{\text{prior}}$: the ground truth reward;
  \item $\hat p$: the marginal distribution in the reverse process induced by guidance $\hat f$ (returned by~\cref{alg:main-alg}).
\end{itemize}
For simplicity, let
\begin{align*}
  \hat v(\xb,t) := & \mathbb{E}_{r\sim \hat R^{\hat n}(\cdot | \xb,t)} \exp( \eta \cdot r )\\
  v^{\star}(\xb,t) := & \mathbb{E}_{r\sim R^{\text{prior}}(\cdot | \xb,t)} \exp( \eta \cdot r )\\
  V(\xb,t) := & \hat v(\xb,t) - v^{\star}(\xb,t).
\end{align*}

Throughout this section, we discuss some natural conditions under which small error from estimating $v^\star(\xb,t)$ leads to small error on estimating $\nabla_{\xb} \ln v^\star(\xb,t)$. The conditions are related to the smoothness of functions $V, v^\star,\hat v$, and the distribution $p$.

Firstly, we present the following lemma, showing that it's sufficient to control $\nabla f_1(\xb) - \nabla f_2(\xb)$:

\begin{lemma}\label{lem:grad-ln-f-to-f}
  Let $f_1,f_2$ be $\RR^d \to \RR$, then
  \begin{align*}
    & \left\|\nabla \ln f_1(\xb) - \nabla \ln f_2(\xb)\right\|_2 \\
    \le & \frac{1}{\inf_\xb |f_1(\xb)|}\|\nabla f_1(\xb) - \nabla f_2(\xb)\|_2
    + \frac{\sup_{\xb}\|\nabla f_2(\xb)\|_2}{\inf_{\xb}|f_1(\xb)|\inf_{\xb}|f_2(\xb)|}\left|f_2(\xb) - f_1(\xb)\right|
  \end{align*}
\end{lemma}
\subsection{Smoothness assumptions on both the classifier and the distribution}\label{sec:smooth-class-dist-assumption}
Our first set of assumptions is based on the smoothness of the functions and the smoothness of the distributions: there exists $M, L, L_p>0$, s.t.
\begin{itemize}
  \item for all $t$, $\EE_{\xb \sim \hat p_t}\left[\left(\sum_{i=1}^d \frac{\partial^2 V(\xb,t)}{\partial x_i^2} \right)^2\right] \le M^2$;
  \item for all $t$, $\sup_{\xb}\|\nabla_{\xb} \hat v(\xb,t)\|_2, \sup_{\xb}\|\nabla_{\xb} v^{\star}(\xb,t)\|_2 \le L$;
  \item for all $t$, $\EE_{\xb \sim p}\left[\|\nabla \log p(\xb)\|_2^2\right] \le L_p^2$.
\end{itemize}
We now present the following results that relates the difference of gradients to difference of function value:
\begin{lemma}\label{lem:f-to-grad-f}
  Let $F(\cdot)$ be function that map from $\RR^d \to \RR$, and $p(\cdot)$ be a PDF over $\RR^d$.
  Suppose $\lim_{x_i \to \infty}F(\xb) \cdot \frac{\partial F(\xb)}{\partial x_i} p(\xb) = 0$ for all $i$, and $\xb_{-i}$. Then $\EE_{\xb \sim p} \|\nabla F(\xb)\|_2^2$ is bounded by:
  \begin{align*}
    \left(\sqrt{\EE_{\xb \sim p}\left[\left(\sum_{i=1}^d \frac{\partial^2 F(\xb)}{\partial x_i^2} \right)^2\right]}
    + \sup_{\xb}\left(\|\nabla F(\xb)\|_2\right)\sqrt{\EE_{\xb \sim p}\left[\|\nabla \log p(\xb)\|_2^2\right]}\right)\sqrt{\EE_{\xb \sim p}\left[F^2(\xb)\right]}.
  \end{align*}
\end{lemma}
Given these, we can show a version of \cref{assum:reward-class-condition}. By definition, for all $t$,
\begin{align*}
  \inf_{\xb}|\hat v(\xb,t)|, \inf_{\xb}|v^{\star}(\xb,t)| \ge \exp(-\eta R_{\max}).
\end{align*}
By Lemma~\ref{lem:grad-ln-f-to-f} and Lemma~\ref{lem:f-to-grad-f}: for all $t$.
\begin{align}
  & \EE_{\xb \sim \hat p_t}\left[\|\nabla_{\xb} \ln \hat v(\xb,t) - \nabla_{\xb} \ln v^{\star}(\xb,t)\|_2^2\right] \nonumber\\
  \le & 2\exp(2\eta R_{\max})\EE_{\xb \sim \hat p_t}\left[\|\nabla_{\xb}  \hat v(\xb,t) - \nabla_{\xb}  v^{\star}(\xb,t)\|_2^2\right]
  + 2\exp(4\eta R_{\max}) L \EE_{\xb \sim \hat p_t}\left[|\hat v(\xb,t) - v^{\star}(\xb,t)|^2\right] \nonumber\\
  \le & 2\exp(2\eta R_{\max})\left(M + L \cdot L_p\right)\sqrt{\EE_{\xb \sim \hat p_t}\left[|\hat v(\xb,t) - v^{\star}(\xb,t)|^2\right]} \nonumber\\
  & + 2\exp(4\eta R_{\max}) L \EE_{\xb \sim \hat p_t}\left[|\hat v(\xb,t) - v^{\star}(\xb,t)|^2\right] \nonumber
\end{align}
This will only change the proof in Appendix~\ref{app:proof} slightly.

\subsection{Smoothness assumption and gradient estimator}
In this section, we introduce another set of conditions based on the gradient estimator introduced in \cite{flaxman2004online}. We require smoothness assumptions on the functions and an additional gradient estimator. For any function $f:\RR^d \to \RR$, we define the gradient estimator $\widehat{\nabla f}$ to be:
\begin{align*}
    \widehat{\nabla f}(\xb) := \frac{d}{\delta} \EE_{\ub \sim \Ucal(S^{d-1})}\left[f(\xb + \delta \ub)\ub\right],
\end{align*}
where $S^{d-1}:= \{\xb \in \RR^d: \|\xb\|_2 = 1\}$ and $\delta >0$ is a free parameter. We use $\Ucal(S^{d-1})$ to denote the uniformly random distribution over $S^{d-1}$.
We assume there exist $M>0$, s.t.:
\begin{itemize}
    \item for all $t$, $\sup_{\xb}\|\nabla_{\xb}^2 \hat v(\xb,t)\|_2, \sup_{\xb}\|\nabla_{\xb}^2 v^{\star}(\xb,t)\|_2 \le M$
\end{itemize}
We first present two properties of the gradient estimator. 
\begin{lemma}\label{lem:grad-est-diff}
    Let $f_1, f_2$ be two functions that map from $\RR^d$ to $\RR$, then
    \begin{align*}
        \|\widehat{\nabla f_1}(\xb) - \widehat{\nabla f_2}(\xb)\|_2  \le \frac{d}{\delta} \EE_{\ub \sim \Ucal(S^{d-1})}\left[\left|f_1(\xb + \delta \ub) - f_2(\xb + \delta \ub)\right|\right].
    \end{align*}
\end{lemma}

\begin{lemma}\label{lem:grad-est-err}
    Let $f$ be $\RR^d \to \RR$ and $\sup_{\xb}\|\nabla^2 f(\xb)\|_2 \le M$, then
    \begin{align*}
        \left\|\widehat{\nabla f}(\xb) - \nabla f(\xb)\right\|_2 \le \frac{d \delta M}{2}.
    \end{align*}
\end{lemma}
Given these, we can show that, for all $t$, 
\begin{align*}
    & \EE_{\xb \sim \hat p_t}\left[\|\nabla_{\xb}  \hat v(\xb,t) - \nabla_{\xb}  v^{\star}(\xb,t)\|_2^2\right] \\
    \le & 3\EE_{\xb \sim \hat p_t}\left[\|\nabla_{\xb}  \hat v(\xb,t) - \widehat{\nabla_{\xb}  \hat v}(\xb,t)\|_2^2\right]
    + 3\EE_{\xb \sim \hat p_t}\left[\|\widehat{\nabla_{\xb}  \hat v}(\xb,t) - \widehat{\nabla_{\xb}  v^{\star}}(\xb,t)\|_2^2\right]\\
    &+ 3\EE_{\xb \sim \hat p_t}\left[\|\widehat{\nabla_{\xb}  v^{\star}}(\xb,t) - \nabla_{\xb}  v^{\star}(\xb,t)\|_2^2\right]\\
    \le & \frac{3d^2 \delta^2 M^2}{2} + \frac{3d^2}{\delta^2}\EE_{\xb \sim \hat p_t}\left[\left(\EE_{\ub \sim \Ucal(S^{d-1})}\left[\hat v(\xb+\delta\ub,t) - v^{\star}(\xb+\delta\ub,t)\right]\right)^2\right] \\
    \le & \frac{3d^2 \delta^2 M^2}{2} + \frac{3d^2}{\delta^2}\EE_{\xb \sim \hat p_t}\left[\left|\EE_{\ub \sim \Ucal(S^{d-1})}\left[\hat v(\xb+\delta\ub,t) - v^{\star}(\xb+\delta\ub,t)\right]\right|\right] \\
    \le & \frac{3d^2 \delta^2 M^2}{2} + \frac{3d^2}{\delta^2}\EE_{\xb \sim \hat p_t,\ub \sim \Ucal(S^{d-1})}\left[\left|\hat v(\xb+\delta\ub,t) - v^{\star}(\xb+\delta\ub,t)\right|\right] \\
    \le & \frac{3d^2 \delta^2 M^2}{2} + \frac{3d^2}{\delta^2}\sqrt{\EE_{\xb \sim \hat p_t,\ub \sim \Ucal(S^{d-1})}\left[\left(\hat v(\xb+\delta\ub,t) - v^{\star}(\xb+\delta\ub,t)\right)^2\right]},
\end{align*}
where for the third inequality, we use the fact that $\hat v(\cdot,\cdot), v^{\star}(\cdot,\cdot) \in [0,1]$. The remaining steps follow from Lemma~\ref{lem:grad-ln-f-to-f}, similar to Section~\ref{sec:smooth-class-dist-assumption}:
\begin{align}
  & \EE_{\xb \sim \hat p_t}\left[\|\nabla_{\xb} \ln \hat v(\xb,t) - \nabla_{\xb} \ln v^{\star}(\xb,t)\|_2^2\right] \nonumber\\
  \le & 2\exp(2\eta R_{\max})\EE_{\xb \sim \hat p_t}\left[\|\nabla_{\xb}  \hat v(\xb,t) - \nabla_{\xb}  v^{\star}(\xb,t)\|_2^2\right]
  + 2\exp(4\eta R_{\max}) L \EE_{\xb \sim \hat p_t}\left[|\hat v(\xb,t) - v^{\star}(\xb,t)|^2\right] \nonumber\\
  \le & 2\exp(2\eta R_{\max})\left(\frac{3d^2 \delta^2 M^2}{2} + \frac{3d^2}{\delta^2}\sqrt{\EE_{\xb \sim \hat p_t,\ub \sim \Ucal(S^{d-1})}\left[\left(\hat v(\xb+\delta\ub,t) - v^{\star}(\xb+\delta\ub,t)\right)^2\right]}\right) \nonumber\\
  & + 2\exp(4\eta R_{\max}) L \EE_{\xb \sim \hat p_t}\left[|\hat v(\xb,t) - v^{\star}(\xb,t)|^2\right] \nonumber
\end{align}

Compared to Algorithm~\ref{alg:main-alg}, this approach requires one to additionally optimize for $\left(\hat v(\xb+\delta\ub,t) - v^{\star}(\xb+\delta\ub,t)\right)^2$, i.e. to make sure $\hat v$ and $v$ are close under some ``wider'' distribution. The value of $\delta$ is a free parameter that can be adjusted to improve the accuracy of the gradient estimator.

\subsection{Proof of Lemma~\ref{lem:grad-ln-f-to-f}}
\begin{align*}
  & \left\|\nabla \ln f_1(\xb) - \nabla \ln f_2(\xb)\right\|_2
  =  \left\|\frac{\nabla f_1(\xb)}{f_1(\xb)} - \frac{\nabla f_2(\xb)}{f_2(\xb)}\right\|_2 \\
  \le & \left\|\frac{\nabla f_1(\xb)}{f_1(\xb)} - \frac{\nabla f_2(\xb)}{f_1(\xb)}\right\|_2
  + \left\|\frac{\nabla f_2(\xb)}{f_1(\xb)} - \frac{\nabla f_2(\xb)}{f_2(\xb)}\right\|_2\\
  = & \frac{\|\nabla f_1(\xb) - \nabla f_2(\xb)\|_2}{|f_1(\xb)|}
  + \frac{1}{|f_1(\xb)f_2(\xb)|}\left\|\nabla f_2(\xb) (f_2(\xb) - f_1(\xb))\right\|_2 \\
  \le & \frac{1}{\inf_\xb |f_1(\xb)|}\|\nabla f_1(\xb) - \nabla f_2(\xb)\|_2
  + \frac{\sup_{\xb}\|\nabla f_2(\xb)\|_2}{\inf_{\xb}|f_1(\xb)|\inf_{\xb}|f_2(\xb)|}\left|f_2(\xb) - f_1(\xb)\right|,
\end{align*}

\subsection{Proof of Lemma~\ref{lem:f-to-grad-f}}
We rewrite the target as componentwise form:
\begin{align*}
  \|\nabla F(\xb)\|_2^2 = \sum_{i=1}^d \left(\frac{\partial F(\xb)}{\partial x_i}\right)^2.
\end{align*}
By applying integration by parts to the $i$-th dimension,
\begin{align*}
  & \int \frac{\partial F(\xb)}{\partial x_i} \cdot \frac{\partial F(\xb)}{\partial x_i} p(\xb) \d \xb \\
  = & \int \left. F(\xb) \cdot \frac{\partial F(\xb)}{\partial x_i} p(\xb) \right|_{x_i = -\infty}^{\infty}\d \xb_{-i}
  -\int F(\xb) \left( \frac{\partial^2 F(\xb)}{\partial x_i^2} p(\xb) + \frac{\partial F(\xb)}{\partial x_i}\frac{\partial p(\xb)}{\partial x_i}  \right) \d \xb \\
  = & -\int F(\xb) \left( \frac{\partial^2 F(\xb)}{\partial x_i^2} p(\xb) + \frac{\partial F(\xb)}{\partial x_i}\frac{\partial \log p(\xb)}{\partial x_i} p(\xb)  \right) \d \xb.
\end{align*}
By summing over all coordinates, we get:
\begin{align*}
  \EE_{\xb \sim p} \|\nabla F(\xb)\|_2^2
  = - \EE_{\xb \sim p}\left[F(\xb)\sum_{i=1}^d \frac{\partial^2 F(\xb)}{\partial x_i^2} \right]
  - \EE_{\xb \sim p}\left[F(\xb) \left\langle \nabla F(\xb), \nabla \log p(\xb)\right\rangle \right]
\end{align*}
By Cauchy-Schwarz inequality:
\begin{align*}
  & \EE_{\xb \sim p} \|\nabla F(\xb)\|_2^2  \\
  \le &  \left(\sqrt{\EE_{\xb \sim p}\left[\left(\sum_{i=1}^d \frac{\partial^2 F(\xb)}{\partial x_i^2} \right)^2\right]}  + \sqrt{\EE_{\xb \sim p}\left[\left(\left\langle \nabla F(\xb), \nabla \log p(\xb)\right\rangle\right)^2\right]}\right) \sqrt{\EE_{\xb \sim p}\left[F^2(\xb)\right]} \\
  \le & \left(\sqrt{\EE_{\xb \sim p}\left[\left(\sum_{i=1}^d \frac{\partial^2 F(\xb)}{\partial x_i^2} \right)^2\right]}
  + \sup_{\xb}\left(\|\nabla F(\xb)\|_2\right)\sqrt{\EE_{\xb \sim p}\left[\|\nabla \log p(\xb)\|_2^2\right]}\right)\sqrt{\EE_{\xb \sim p}\left[F^2(\xb)\right]}
\end{align*}

\subsection{Proof of Lemma~\ref{lem:grad-est-diff}}
By Jensen's inequality,
\begin{align*}
    \|\widehat{\nabla f_1}(\xb) - \widehat{\nabla f_2}(\xb)\|_2 
    = & \frac{d}{\delta} \left\|\EE_{\ub \sim \Ucal(S^{d-1})}\left[f_1(\xb + \delta \ub)\ub\right] - \EE_{\ub \sim \Ucal(S^{d-1})}\left[f_2(\xb + \delta \ub)\ub\right]\right\|_2 \\
    = & \frac{d}{\delta} \left\|\EE_{\ub \sim \Ucal(S^{d-1})}\left[\left(f_1(\xb + \delta \ub) - f_2(\xb + \delta \ub)\right)\ub\right]\right\|_2 \\
    \le  & \frac{d}{\delta} \EE_{\ub \sim \Ucal(S^{d-1})}\left[\left\|\left(f_1(\xb + \delta \ub) - f_2(\xb + \delta \ub)\right)\ub\right\|_2\right] \\
    =  & \frac{d}{\delta} \EE_{\ub \sim \Ucal(S^{d-1})}\left[\left|f_1(\xb + \delta \ub) - f_2(\xb + \delta \ub)\right|\left\|\ub\right\|_2\right] \\
    =  & \frac{d}{\delta} \EE_{\ub \sim \Ucal(S^{d-1})}\left[\left|f_1(\xb + \delta \ub) - f_2(\xb + \delta \ub)\right|\right].
\end{align*}

\subsection{Proof of Lemma~\ref{lem:grad-est-err}}
By Taylor's theorem, for all $\xb,\vb$, there exists $\xi(\xb,\vb)$, s.t. 
\begin{align*}
    f(\xb + \vb) = f(\xb) + \langle \nabla f(\xb), \vb\rangle + \frac{1}{2} \vb^\top \nabla^2 f(\xi(\xb,\vb)) \vb.
\end{align*}
Then
\begin{align*}
    & \left\|\widehat{\nabla f}(\xb) - \nabla f(\xb)\right\|_2
    = \left\|\frac{d}{\delta} \EE_{\ub \sim \Ucal(S^{d-1})}\left[f(\xb + \delta \ub)\ub\right] - \nabla f(\xb)\right\|_2 \\
    = & \left\|\frac{d}{\delta} \EE_{\ub \sim \Ucal(S^{d-1})}\left[f(\xb)\ub + \delta \langle \nabla f(\xb), \ub\rangle \ub + \frac{\delta^2}{2} \left(\ub^\top \nabla^2 f(\xi(\xb,\delta\ub)) \ub\right) \ub\right] - \nabla f(\xb)\right\|_2 \\
\end{align*}
Because $\EE_{\ub \sim \Ucal(S^{d-1})}[\ub]=0$, 
\begin{align*}
    \EE_{\ub \sim \Ucal(S^{d-1})}\left[f(\xb)\ub\right] = 0.
\end{align*}
Because $\EE_{\ub \sim \Ucal(S^{d-1})}[\ub\ub^\top]=\frac{1}{d}I$,
\begin{align*}
    \EE_{\ub \sim \Ucal(S^{d-1})}\left[\langle \nabla f(\xb), \ub\rangle \ub\right]
    = 
    \EE_{\ub \sim \Ucal(S^{d-1})}\left[\ub\ub^{\top} \nabla f(\xb) \right]
    =
    \EE_{\ub \sim \Ucal(S^{d-1})}\left[\ub\ub^{\top}\right] \nabla f(\xb) 
    = \frac{1}{d} \nabla f(\xb).
\end{align*}
Thus, by Jensen's inequality, we have:
\begin{align*}
    & \left\|\widehat{\nabla f}(\xb) - \nabla f(\xb)\right\|_2 
    =  \left\|\frac{d}{\delta} \EE_{\ub \sim \Ucal(S^{d-1})}\left[\frac{\delta^2}{2} \left(\ub^\top \nabla^2 f(\xi(\xb,\delta\ub)) \ub\right) \ub\right]\right\|_2 \\
    \le & \frac{d\delta}{2} \EE_{\ub \sim \Ucal(S^{d-1})}\left[\left\| \left(\ub^\top \nabla^2 f(\xi(\xb,\delta\ub)) \ub\right) \ub\right\|_2\right] 
    =  \frac{d\delta}{2} \EE_{\ub \sim \Ucal(S^{d-1})}\left[ \left|\ub^\top \nabla^2 f(\xi(\xb,\delta\ub)) \ub\right| \left\|\ub\right\|_2\right] \\
    = & \frac{d\delta}{2} \EE_{\ub \sim \Ucal(S^{d-1})}\left[ \left|\ub^\top \nabla^2 f(\xi(\xb,\delta\ub)) \ub\right|\right] 
    \le  \frac{d\delta}{2} \EE_{\ub \sim \Ucal(S^{d-1})}\left[ \|\ub\|_2 \|\nabla^2 f(\xi(\xb,\delta\ub)) \ub\|_2\right] \\
    \le & \frac{d\delta M}{2}.
\end{align*}

\section{Proof of Main Theorem}\label{app:proof}
\begin{proof}
  By Assumption~\ref{assum:no-regret},
  \begin{align*}
    \frac{1}{NM}\sum_{i=1}^N \sum_{(t,\xb_t,r)\in D^i}\ln\frac{1}{\hat R^i(r|\xb_t,t)}
    \le & \min_R \frac{1}{NM}\sum_{i=1}^N \sum_{(t,\xb_t,r)\in D^i}\ln\frac{1}{R(r|\xb_t,t)} + \gamma_N\\
    \le & \frac{1}{NM}\sum_{i=1}^N \sum_{(t,\xb_t,r)\in D^i}\ln\frac{1}{R^{\text{prior}}(r|\xb_t,t)} + \gamma_N.
  \end{align*}
  After rearranging, we get
  \begin{align*}
    \frac{1}{NM}\sum_{i=1}^N \sum_{(t,\xb_t,r)\in D^i}\ln\frac{R^{\text{prior}}(r|\xb_t,t)}{\hat R^i(r|\xb_t,t)}
    \le & \gamma_N.
  \end{align*}
  According to \cref{eq:data-collection}, each $(t,\xb_t)$, $r$ is sampled from $R^{\text{prior}}(\cdot | \xb_t,t)$. By taking expectation over $r | \xb_t,t$ for all $\xb_t,t$, we get:
  \begin{align*}
    \frac{1}{NM}\sum_{i=1}^N \sum_{(t,\xb_t)\in D^i}\KL\left(R^{\text{prior}}(\cdot|\xb_t,t)\|\hat R^i(\cdot|\xb_t,t)\right)
    \le & \gamma_N.
  \end{align*}
  By taking expectation over $\xb_t$ and $t$, we get:
  \begin{align*}
    \frac{1}{N}\sum_{i=1}^N \EE_{t\sim \Ucal(t), \xb_t\sim P^{\fb^{i}}_{0\to t}(\cdot | \Ncal(0,I))}\left[\KL\left(R^{\text{prior}}(\cdot|\xb_t,t)\|\hat R^i(\cdot|\xb_t,t)\right)\right]
    \le & \gamma_N,
  \end{align*}
  where $\xb_t$ is sampled from the SDE induced by $\hat R^i$.
  By Pinsker's inequality,
  \begin{align*}
    & \frac{1}{N}\sum_{i=1}^N \EE_{t\sim \Ucal(t), \xb_t\sim P^{\fb^{i}}_{0\to t}(\cdot | \Ncal(0,I))}\left[\left|\TV\left(R^{\text{prior}}(\cdot|\xb_t,t),\hat R^i(\cdot|\xb_t,t)\right)\right|^2\right]\\
    \le &
    \frac{1}{N}\sum_{i=1}^N \frac{1}{2}\EE_{t\sim \Ucal(t), \xb_t\sim P^{\fb^{i}}_{0\to t}(\cdot | \Ncal(0,I))}\left[\KL\left(R^{\text{prior}}(\cdot|\xb_t,t)\|\hat R^i(\cdot|\xb_t,t)\right)\right]
    \le \frac{1}{2}\gamma_N.
  \end{align*}
  This means, there exists $\hat n \in \{1,\ldots, N\}$, s.t.
  \begin{align*}
    \EE_{t\sim\Ucal(T),x\sim P^{\fb^{\hat n}}_{0\to t}(\cdot | \Ncal(0,I))}\left|\TV\left(R^{\text{prior}}(\cdot|\xb_t,t),\hat R^{\hat n}(\cdot|\xb_t,t)\right)\right|^2
    \le & \frac{1}{2}\gamma_N.
  \end{align*}
  Let
  \begin{align*}
    \hat\fb(\xb_t,t) := & \fb^{\hat n}(\xb_t,t) =  \nabla_{\xb_t}\ln \mathbb{E}_{r\sim \hat R^{\hat n}(\cdot | \xb_t,t)} \exp( \eta \cdot r )\\
    \fb^{\star}(\xb_t,t) := &  \nabla_{\xb_t}\ln p(y=1 | \xb_t)=\nabla_{\xb_t}\ln \mathbb{E}_{r\sim R^{\text{prior}}(\cdot | \xb_t,t)} \exp( \eta \cdot r )
  \end{align*}
  and
  \begin{align*}
    \hat p_t := P^{\hat\fb}_{0\to t}(\cdot | \Ncal(0,I)),
    \quad
    p_t := P^{\fb^{\star}}_{0\to t}(\cdot | q_T),
  \end{align*}
  recall that $q_T$ is the marginal distribution of the forward process at time $T$.
  By Lemma~\ref{lem:sde-kl-ode},
  \begin{align*}
    \frac{\partial}{\partial t}\KL(\hat p_t \| p_t)
    = & -g^2(T-t)\EE_{x\sim \hat p_t}\left[\left\|\nabla\log\frac{\hat p_t(x)}{p_t(x)}\right\|_2^2\right]\\
    &  + \EE_{x\sim \hat p_t}\left[\left\langle g^2(T-t)\left(\hat\fb(x,t) - \fb^{\star}(x,t) \right), \nabla\log\frac{\hat p_t(x)}{p_t(x)}\right\rangle\right] \\
    \le & \frac{1}{4}g^2(T-t)\EE_{x\sim \hat p_t}\left[\left\|\hat\fb(x,t) - \fb^{\star}(x,t) \right\|_2^2\right].
  \end{align*}
  By integrating over $t \in [0,T]$, we get:
  \begin{align*}
    \KL(\hat p_T \| p_T) \le & \KL(\hat p_0 \| p_0) + \frac{1}{4}\int_0^Tg^2(T-t)\EE_{x\sim \hat p_t}\left[\left\|\hat\fb(x,t) - \fb^{\star}(x,t) \right\|_2^2\right] \d t,
  \end{align*}
  where $\hat p_0 = \Ncal(0,I)$ and $p_0 = q_T$.
  By Assumption~\ref{assum:convergence-forward}, $\KL(\hat p_0 \| p_0) \le \epsilon_T$.
  By Assumption~\ref{assum:reward-class-condition},
  \begin{align*}
    & \int_0^Tg^2(T-t)\EE_{x\sim \hat p_t}\left[\left\|\hat\fb(x,t) - \fb^{\star}(x,t) \right\|_2^2\right] \d t \\
    = & T\|g\|_{\infty}^2\EE_{t\sim\Ucal(T),x\sim \hat p_t}\left[\left\|\hat\fb(x,t) - \fb^{\star}(x,t) \right\|_2^2\right]  \\
    \le & T\|g\|_{\infty}^2L^2\EE_{t\sim\Ucal(T),x\sim \hat p_t}\left[\left|\mathbb{E}_{r\sim \hat R^{\hat n}(\cdot | \xb_t,t)} \exp( \eta \cdot r ) - \mathbb{E}_{r\sim R^{\text{prior}}(\cdot | \xb_t,t)} \exp( \eta \cdot r ) \right|^2\right]  \\
    \le & T\|g\|_{\infty}^2L^2\EE_{t\sim\Ucal(T),x\sim \hat p_t}\left[\left|\TV(\hat R^{\hat n}(\cdot | \xb_t,t),R^{\text{prior}}(\cdot | \xb_t,t)) \right|^2\right]  \\
    \le & \frac{1}{2}T\|g\|_{\infty}^2L^2\gamma_N
  \end{align*}
  To conclude:
  \begin{align*}
    \KL(\hat p_T \| p_T) \le \epsilon_T + \frac{1}{2}T\|g\|_{\infty}^2L^2\gamma_N.
  \end{align*}

\end{proof}

\section{Technical Lemmas}
\label{sec:technical-lemmas}
The following lemma shows that the finite-dimensional distribution of the forward and reverse SDEs are the same.
\begin{lemma}[Section 5 of~\cite{anderson1982reverse}]\label{lem:joint-dist-reverse-forward-sde}
Let $\{\bar\xb_{\tau}\}_{\tau}$ be the process generated by the forward SDE in \cref{eq:forward-sde}, and $\{\xb_{t}\}_{t}$ be the process generated by the reverse SDE in \cref{eq:reverse-sde}. Then for all $s > t$, the conditional distribution $\bar \xb_t | \bar \xb_s$ and $\xb_{T-t} | \bar \xb_{T-s}$ have the same density, i.e. for all $x,x' \in \RR^d$, 
\begin{equation*}
    \Pr\left[ \bar \xb_t = x' | \bar \xb_s = x \right] = \Pr\left[\xb_{T-t} = x'|  \xb_{T-s} = x\right].
\end{equation*}
\end{lemma}
This result is proved by considering the backward Kolmogorov equation of the forward process. The proof is presented in Section 5 of~\cite{anderson1982reverse}. For completeness, we include the proof in Section~\ref{sec:proof-anderson}.

This result implies that the reverse and the forward SDE have the same ``joint distribution'' (in the sense of \emph{finite-dimensional distribution}).
This can be proved by two steps:
\begin{enumerate}
    \item using the Fokker-Planck equation to show that the forward and reverse SDE have the same marginal distribution given proper initial conditions;
    \item write the finite-dimensional distribution as the product of a sequence of conditional distributions and the initial marginal distribution by using Markovian property iteratively.
\end{enumerate}

The following lemma upper bound the KL-divergence between the marginal distributions of two SDEs in terms of the difference in their drift terms:
\begin{lemma}[Lemma 6 of \cite{chen2023improved}]\label{lem:sde-kl-ode}
  Consider the two Ito processes:
  \begin{align*}
    dX_t =& F_1(X_t, t ) \d t + g(t) \d w_t \quad X_0 = a\\
    dY_t =& F_2(Y_t, t ) \d t + g(t) \d w_t \quad Y_0 = a
  \end{align*}
  wher $F_1,F_2,g$ are continuous functions and may depend on $a$. We assume the uniqueness and regularity conditions hold:
  \begin{enumerate}
    \item The two SDEs have unique solutions.
    \item $X_t,Y_t$ admit densities $p^1_t, p^2_t \in C^2(\mathbb{R}^d)$ for $t>0$
  \end{enumerate}

  And define the Fisher information information between $p_t,q_t$ as:
  \begin{align*}
    J(p_t \| q_t) := \int p_t (x) \norm{\nabla \log \frac{p_t(x)}{q_t(x)}}^2 \d x
  \end{align*}

  Then for any $t>0$, the evolution of $\KL(p^1_t \| p^2_t)$ is given by:
  \begin{align*}
    \frac{\partial}{\partial t} \textit{\text{KL}}(p^1_t \| p^2_t) = -g(t)^2 J(p^1_t \| p^2_t) + \EE_{x\sim p^1_t}\left[\left\langle F_1(X_t,t) - F_2(X_t,t), \nabla \log \frac{p^1_t(x)}{p^2_t(x)} \right\rangle\right]
  \end{align*}

\end{lemma}

\subsection{Proof of Lemma~\ref{lem:joint-dist-reverse-forward-sde}}\label{sec:proof-anderson}
Let $q(x_s,s | x_t, t)$ be the conditional distribution of~\cref{eq:forward-sde}, where $s\ge t$ is the time index. And $q(x_t,t)$ be the marginal distribution at time $t$.
By the backward Kolmogorov equation:
\begin{equation}\label{eq:kbe}
  \frac{\partial q(\bar x_s,s | \bar x_t, t)}{\partial t} 
  = 
  -\sum_i \hb^i(\bar x_t,t) \frac{\partial q(\bar x_s, s | \bar x_t, t)}{\partial \bar x^i_t} 
  - \frac12 g^2(t)\sum_{i} \frac{\partial^2 q(\bar x_s, s | \bar x_t, t)}{\partial  \left(\bar x^i_t\right)^2}
\end{equation}
According to Fokker-Planck equation, the marginal distribution $q(\bar x_t,t)$ satisfies:
\begin{equation}\label{eq:fpe}
\frac{\partial q(\bar x_t, t)}{\partial t} 
= 
- \sum_i \frac{\partial}{\partial \bar x^i_t} \left[ q(\bar x_t, t) \hb^i(\bar x_t,t) \right] 
+ \frac12 g^2(t)\sum_{i} \frac{\partial^2  q(\bar x_t,t) }{\partial \left(\bar x^i_t\right)^2} 
\end{equation}
Because the joint distribution satisfies:
\begin{align*}
  q(\bar x_s,s, \bar x_t, t) = q(\bar x_s,s | \bar x_t, t)q(\bar x_t,t),
\end{align*}
we have 
\begin{align*}
    \frac{\partial q(\bar x_s,s, \bar x_t, t)}{\partial t}
    = & q(\bar x_t,t)\frac{\partial q(\bar x_s,s | \bar x_t, t)}{\partial t} + q(\bar x_s,s | \bar x_t, t)\frac{\partial q(\bar x_t,t)}{\partial t}.
\end{align*}
Plug in \cref{eq:kbe} and \cref{eq:fpe}, we have:
\begin{align*}
    & \frac{\partial q(\bar x_s,s, \bar x_t, t)}{\partial t}\\
    = & 
    q(\bar x_t,t)\left(  -\sum_i \hb^i(\bar x_t,t) \frac{\partial q(\bar x_s, s | \bar x_t, t)}{\partial \bar x^i_t} 
  - \frac12 g^2(t)\sum_{i} \frac{\partial^2 q(\bar x_s, s | \bar x_t, t)}{\partial  \left(\bar x^i_t\right)^2}\right) \\
  & + q(\bar x_s,s | \bar x_t, t) 
  \left(- \sum_i \frac{\partial}{\partial \bar x^i_t} \left[ q(\bar x_t, t) \hb^i(\bar x_t,t) \right] 
+ \frac12 g^2(t)\sum_{i} \frac{\partial^2  q(\bar x_t,t) }{\partial \left(\bar x^i_t\right)^2} \right).
\end{align*}
For the first and third terms:
\begin{align*}
    & - q(\bar x_t,t) \sum_i \hb^i(\bar x_t,t) \frac{\partial q(\bar x_s, s | \bar x_t, t)}{\partial \bar x^i_t}
    - q(\bar x_s,s | \bar x_t, t) \sum_i \frac{\partial}{\partial \bar x^i_t} \left[ q(\bar x_t, t) \hb^i(\bar x_t,t) \right] \\ 
    = & 
    -  \sum_i \hb^i(\bar x_t,t) \left(q(\bar x_t,t)\frac{\partial q(\bar x_s, s | \bar x_t, t)}{\partial \bar x^i_t} + q(\bar x_s,s | \bar x_t, t)\frac{\partial}{\partial \bar x^i_t}q(\bar x_t, t)\right) \\
    &
    - q(\bar x_s,s | \bar x_t, t) q(\bar x_t, t)\sum_i \frac{\partial}{\partial \bar x^i_t} \hb^i(\bar x_t,t) \\
    = & 
    -  \sum_i \hb^i(\bar x_t,t) \frac{\partial q(\bar x_s, s, \bar x_t, t)}{\partial \bar x^i_t}
    - q(\bar x_s,s, \bar x_t, t) \sum_i \frac{\partial}{\partial \bar x^i_t} \hb^i(\bar x_t,t)   \\
    = & 
    -  \sum_i \frac{\partial}{\partial \bar x^i_t}\left[\hb^i(\bar x_t,t) q(\bar x_s, s, \bar x_t, t)\right].
\end{align*}
For the second and forth terms:
\begin{align*}
    & \frac12 g^2(t)\sum_{i}\left(-q(\bar x_t,t)\frac{\partial^2 q(\bar x_s, s | \bar x_t, t)}{\partial  \left(\bar x^i_t\right)^2} + q(\bar x_s,s | \bar x_t, t) 
  \frac{\partial^2  q(\bar x_t,t) }{\partial \left(\bar x^i_t\right)^2} \right)\\
  = & 
  -\frac12 g^2(t)\sum_{i}\left(
  q(\bar x_s, s| \bar x_t, t)\frac{\partial^2 q(\bar x_t, t)}{\partial  \left(\bar x^i_t\right)^2}
  + 
  2\frac{\partial q(\bar x_s, s| \bar x_t, t)}{\partial  \bar x^i_t}
  \frac{\partial q(\bar x_t, t)}{\partial  \bar x^i_t}
  + 
  q(\bar x_t, t)\frac{\partial^2 q(\bar x_s, s| \bar x_t, t)}{\partial  \left(\bar x^i_t\right)^2}\right)\\
  &+ 
  \frac12 g^2(t)\sum_{i}\left(2q(\bar x_s, s| \bar x_t, t)\frac{\partial^2 q(\bar x_t, t)}{\partial  \left(\bar x^i_t\right)^2} + 2\frac{\partial q(\bar x_s, s| \bar x_t, t)}{\partial  \bar x^i_t}
  \frac{\partial q(\bar x_t, t)}{\partial  \bar x^i_t} \right) \\
  = & 
  -\frac12 g^2(t)\sum_{i} \frac{\partial^2 q(\bar x_s, s, \bar x_t, t)}{\partial \left(\bar x^i_t\right)^2} 
   + g^2(t)\sum_{i}\left(q(\bar x_s, s| \bar x_t, t)\frac{\partial^2 q(\bar x_t, t)}{\partial  \left(\bar x^i_t\right)^2} + \frac{\partial q(\bar x_s, s| \bar x_t, t)}{\partial  \bar x^i_t}
  \frac{\partial q(\bar x_t, t)}{\partial  \bar x^i_t} \right) \\
  = & 
  -\frac12 g^2(t)\sum_{i} \frac{\partial^2 q(\bar x_s, s, \bar x_t, t)}{\partial \left(\bar x^i_t\right)^2} 
   + g^2(t)\sum_{i}\frac{\partial}{\partial  \bar x^i_t}\left[q(\bar x_s, s| \bar x_t, t)
  \frac{\partial q(\bar x_t, t)}{\partial  \bar x^i_t} \right].
\end{align*}
To summarize, the joint distribution satisfies the following PDE 
\begin{align*}
  \frac{\partial q(\bar x_s,s, \bar x_t, t)}{\partial t} = -\sum_i \frac{\partial}{\partial \bar x^i_t} \left[ q(\bar x_s,s, \bar x_t, t) \bar\hb^i (\bar x_t,t) \right] - \frac12 g^2(t)\sum_{i} \frac{\partial^2 q(\bar x_s,s, \bar x_t, t)}{\partial \left(\bar x^i_t\right)^2},
\end{align*}

where $\bar\hb^i (\bar x_t,t)$ is defined as:
\begin{align*}
  \bar\hb^i (\bar x_t,t) := \hb^i(\bar x_t,t) - \frac{1}{q(\bar x_t,t)} g^2(t)\frac{\partial}{\partial \bar x^i_t} q(\bar x_t,t).
\end{align*}

Divide $q(\bar x_s,s)$ on both sides, we get the following PDE for the conditional distribution $q(\bar x_t, t | \bar x_s,s)$ (notice that this one is conditioning on the future):
\begin{equation}\label{eq:future-condition-sde}
  \frac{\partial q(\bar x_t,t | \bar x_s, s)}{\partial t} = -\sum_{i} \frac{\partial}{\partial \bar x^i_t} \left[ q(\bar x_t, t | \bar x_s,s) \bar\hb^i(\bar x_t,t) \right] - \frac12 g^2(t)\sum_{i} \frac{\partial^2 q(\bar x_t, t | \bar x_s,s)}{\partial \left(\bar x^i_t\right)^2}
\end{equation}

Recall that the reverse process is defined by:
\begin{equation*}
  \d \xb = \left[-\hb(\xb,T-t) + g^2(T-t) \nabla \log q_{T-t}(\xb)\right]\d t + g(T-t) \d \wb, \quad t \in [0,T].
\end{equation*}
The Fokker-Planck equation is given by:
\begin{equation*}
\frac{\partial p(x_t, t)}{\partial t} 
= 
- \sum_i \frac{\partial}{\partial x^i_t} \left[ -p(x_t, t) \bar\hb^i(x_t,T-t) \right] 
+ \frac12 g^2(T-t)\sum_{i} \frac{\partial^2  p(x_t,t) }{\partial \left(x^i_t\right)^2}. 
\end{equation*}
We substitute $t$ with $T-t$ and get:
\begin{equation*}
-\frac{\partial p(x_{T-t}, T-t)}{\partial t} 
= 
\sum_i \frac{\partial}{\partial x^i_{T-t}} \left[ p(x_{T-t}, T-t) \bar\hb^i(x_{T-t},t) \right] 
+ \frac12 g^2(t)\sum_{i} \frac{\partial^2  p(x_{T-t},T-t) }{\partial \left(x^i_{T-t}\right)^2}. 
\end{equation*}
This PDE is identical to \cref{eq:future-condition-sde}. By choosing the proper initial condition, we finish the proof.

\section{Additional Details of Training and Evaluation}
\label{sec:training}
We provide more information about the experimental setup and then overview some of the details of training and evaluation of \ALG{} in this section. The experiments are split up into two parts: image tasks (image compression and aesthetic evaluation) and biological sequence tasks (5' untranslated regions and DNA enhancer sequences).

All experiments were conducted on a single NVIDIA A6000 GPU.

Our classifier network is constructed to predict a histogram over the reward space and trained with cross-entropy loss. When training the classifier for the first iteration of DAgger (without any guidance) we label all states in the trajectory with the reward of the final state. For subsequent iterations, we only use states that are part of the \textit{rollout} (i.e. the states in the trajectory that are computed solely using the prior, after the \textit{rollin} section which uses the latest classifier).

To apply classifier guidance, we compute the empirical expectation of the classifier. That is for $B$ buckets ($r_1, \ldots, r_B$) and predictor $\hat R^n(\cdot | \xb_t, t)$, we compute:
\begin{align*}
  f^n(x_t, t) = \nabla_{\xb_t} \ln \left( \sum_{i=1}^B \hat R^n(r_i | \xb_t, t) \exp( \eta \cdot r_i ) \right).
\end{align*}

Because the gradient is invariant to the reward scale, we set $r_i$ to be a $B$ equally spaced partition of $[0,1]$.

\subsection{Experiment Task Details}

\subsubsection{Comparison Tasks}

\textbf{Image Compression.}
The reward is the negative file size of the generated image, a non-differentiable compression score. We use Stable Diffusion~v1.5 \citep{rombach2022stable} as the base model for this continuous-time diffusion task.

\textbf{Image Aesthetics.}
Here the objective is to maximize the LAION aesthetic score \citep{schumman2022}, obtained from a CLIP encoder followed by an MLP trained on human 1-to-10 ratings. This benchmark is standard in image-generation studies \citep{black2024ddpo,domingoenrich2025adjoint,li2024derivative,uehara2024finetuning,uehara2024feedback}. We again employ Stable Diffusion~v1.5 as the base model.

\textbf{5\,$'$ Untranslated Regions (5\,$'$ UTR) and DNA Enhancers.}
For sequence generation we aim to maximize the mean ribosome load (MRL) of 5\,$'$ UTRs measured via polysome profiling \citep{sample2019human}. Following \citet{li2024derivative}, the base model is that of \citet{sahoo2024simple} trained on \cite{sample2019human}. Likewise, using the same base model, we optimize enhancer sequences according to expression levels predicted by the Enformer model \citep{avsec2021effective} in the HepG2 cell line.

\subsubsection{Comparison Methods}

\textbf{Best-of-$N$.}
We draw \(N\) independent samples from the base diffusion model and retain the single sample with the highest reward.

\textbf{DPS.}
Diffusion Posterior Sampling (DPS) is a training-free variant of classifier guidance originally proposed for continuous diffusion models \citep{chung2023diffusion} and subsequently adapted to discrete diffusion \citep{li2024derivative}. We use the state-of-the-art implementation of \citet{nisonoff2025unlocking}.

\textbf{SMC.}
Sequential Monte Carlo (SMC) methods \citep{del2014particle,wu2023practical, trippe2022diffusion} are a class of methods which use importance sampling on a number of rollouts, and select the best sample. However, note that SMC based methods do this across the \textit{entire} batch, not at a per sample level.

\textbf{SVDD-MC.}
SVDD-MC \citep{li2024derivative} evaluates the expected reward of \(N\) candidates from the base model under an estimated value function and selects the candidate with the highest predicted return.

\textbf{SVDD-PM.}
SVDD-PM \citep{li2024derivative} is similar to SVDD-MC except that it uses the true reward for each candidate instead of relying on value-function estimates.

\subsection{Image Task Details (Image Compression and Aesthetic)}
For both image tasks, we use a lightweight classifier network, with model checkpoints sized at 4.93MB (compression) and 10.60MB (aesthetic).

The prompts are generated under the configuration of SVDD~\cite{li2024derivative}. In the image compression task, we generate 1,400 images per iteration and train for one epoch. The results are reported using the checkpoint from the 4th iteration.
For the aesthetic evaluation task, we generate 10,500 images in the first iteration, followed by 1,400 images in each subsequent iteration, and train for 6 epochs. The results are reported using the checkpoint from the 8th iteration. For the compression task, we adopt the same architecture as~\citet{li2024derivative}. For the aesthetic task, we add five additional residual layers to account for its increased complexity. Notably, our classifier takes the latent representation as direct input, without reusing the VAE (as in SVDD for compression), CLIP (SVDD for aesthetics), or the U-Net (SVDD-PM for estimating $\hat{x}_0$). This results in a significantly more lightweight design, both in network size (approximately 10MB compared to over 4GB) and in runtime (\cref{fig:inference_time}).

\subsubsection{Hyperparameters}
For the image tasks, the following hyperparameters are used:
\begin{table}[H]
  \centering
    \caption{Image Task Hyperparameters. If two values are provided, the first corresponds to the compression task, and the second to the aesthetic task.}
  \begin{tabular}{ll}
    \toprule
    \textbf{Hyperparameter}                & \textbf{Value} \\
    \midrule
    Seed                                   & 43 \\
    Learning rate                          & $1\times10^{-4}$ \\
    Optimizer betas                        & (0.9, 0.999) \\
    Weight decay                           & 0.0 \\
    Gradient accumulation steps            & 1 \\
    Batch size (classifier)                & 8 \\
    Batch size (inference)                 & 1 \\
    Guidance scale                         & 150/75 \\
    Train iterations (per round)           & 1/6 \\
    Eval interval (epoches)                  & 1 \\
    \bottomrule
  \end{tabular}
  \label{tab:hyperparams_img}
\end{table}
\subsection{Sequence Task Details (5' UTR and DNA Enhancer)}
Following \citet{li2024derivative}, for the Enhancer task, we use an Enformer model \cite{avsec2021effective}, and for the 5' UTR task, we use ConvGRU model \cite{dey2017gatevariants}. We reference our classifier guidance implementation based off of \cite{li2024derivative}. Both of these tasks are discrete diffusion problems and use the masking setup of \citet{sahoo2024simple}, following \citet{li2024derivative}.

We use the following hyperparameters for the sequence tasks. In order to not be bound to a specific $\eta$, we take use the gradient of the empirical mean computing the samples for the next iteration with a guidance scale of 10. We can then change $\eta$ at test time as shown in the paper. Due to the more light-weight nature of these tasks, we did 8 iterations of training for the DNA enhancer task. Likewise, we did 7
iterations for the 5' UTR task. Unlike the image tasks, we do not reinitalize the classifier network for each iteration.

\subsubsection{Hyperparameters}

For the DNA enhancer and 5' UTR tasks, the following hyperparameters are used:
\begin{table}[H]
  \centering
    \caption{DNA Enhancer and 5' UTR Task Hyperparameters}
  \begin{tabular}{ll}
    \toprule
    \textbf{Hyperparameter}                & \textbf{Value} \\
    \midrule
    Seed                                   & 43 \\
    Learning rate                          & $1\times10^{-4}$ \\
    Optimizer betas                        & (0.9, 0.95) \\
    Weight decay                           & 0.01 \\
    Gradient accumulation steps            & 4 \\
    Batch size (classifier)                & 5 \\
    Batch size (inference)                 & 20 \\
    Guidance scale                         & 10 \\
    Train iterations (per round)           & 200 \\
    Initial train iterations               & 600 \\
    Classifier epochs                      & 1 \\
    Eval interval (steps)                  & 20 \\
    \bottomrule
  \end{tabular}

  \label{tab:hyperparams}
\end{table}

\section{More Image Samples}
\label{sec:image-samples}

\begin{figure}[ht]
  \centering
  \includegraphics[width=\textwidth]{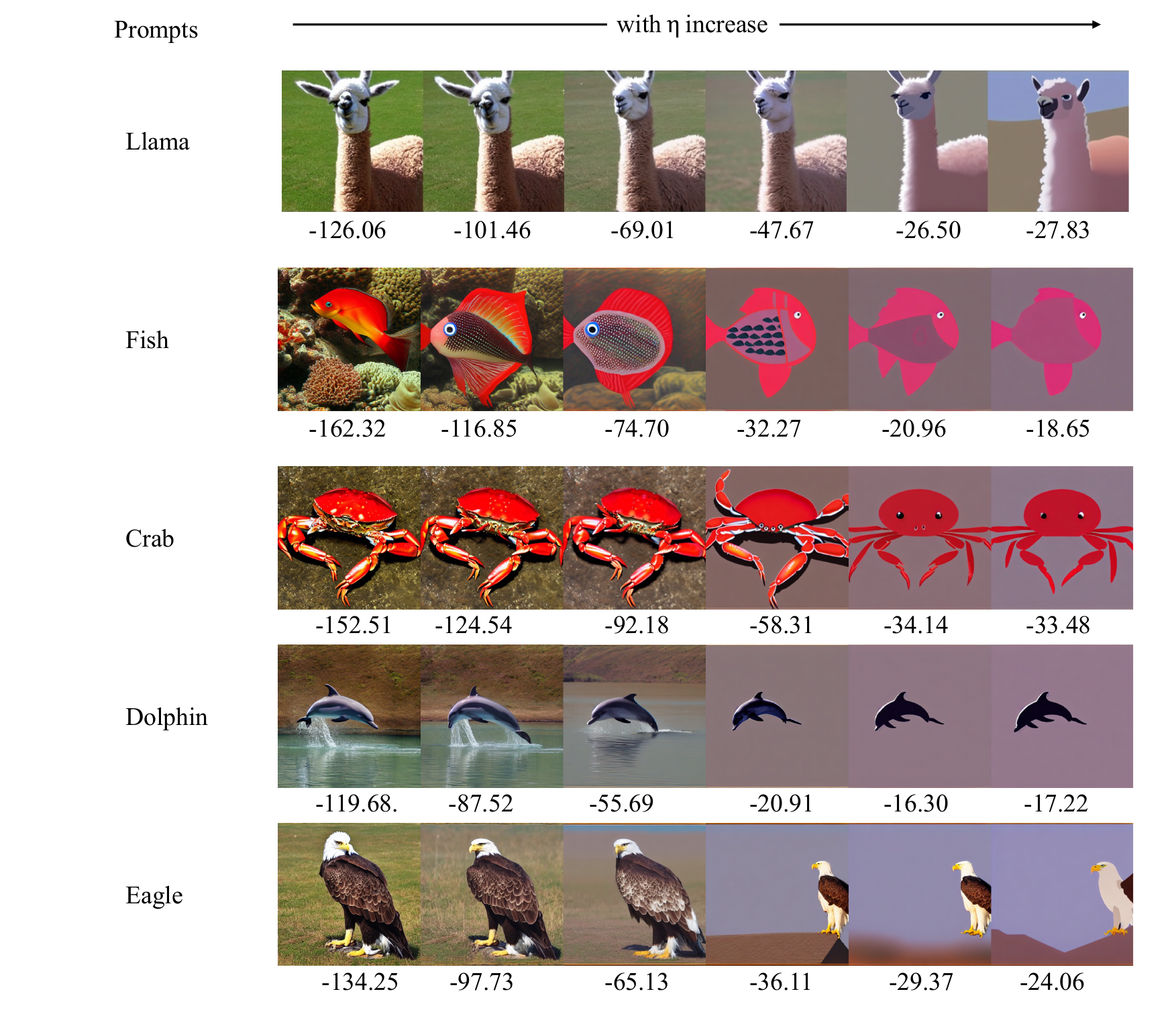}
  \vspace{-0.2in}
  \caption{Additional images generated by \ALG{} with varying $\eta$ values and their corresponding rewards on the image compression task. The prompts used were not seen during training, demonstrating the generalization capability of our method.}
  \label{fig:more different comp images}
\end{figure}

\cref{fig:more different comp images} provides further results for the image compression task using prompts that were not seen during training. The consistent performance across these novel inputs demonstrates the generalization capability of \ALG{}.

\begin{figure}[ht]
  \centering
  \includegraphics[width=\textwidth]{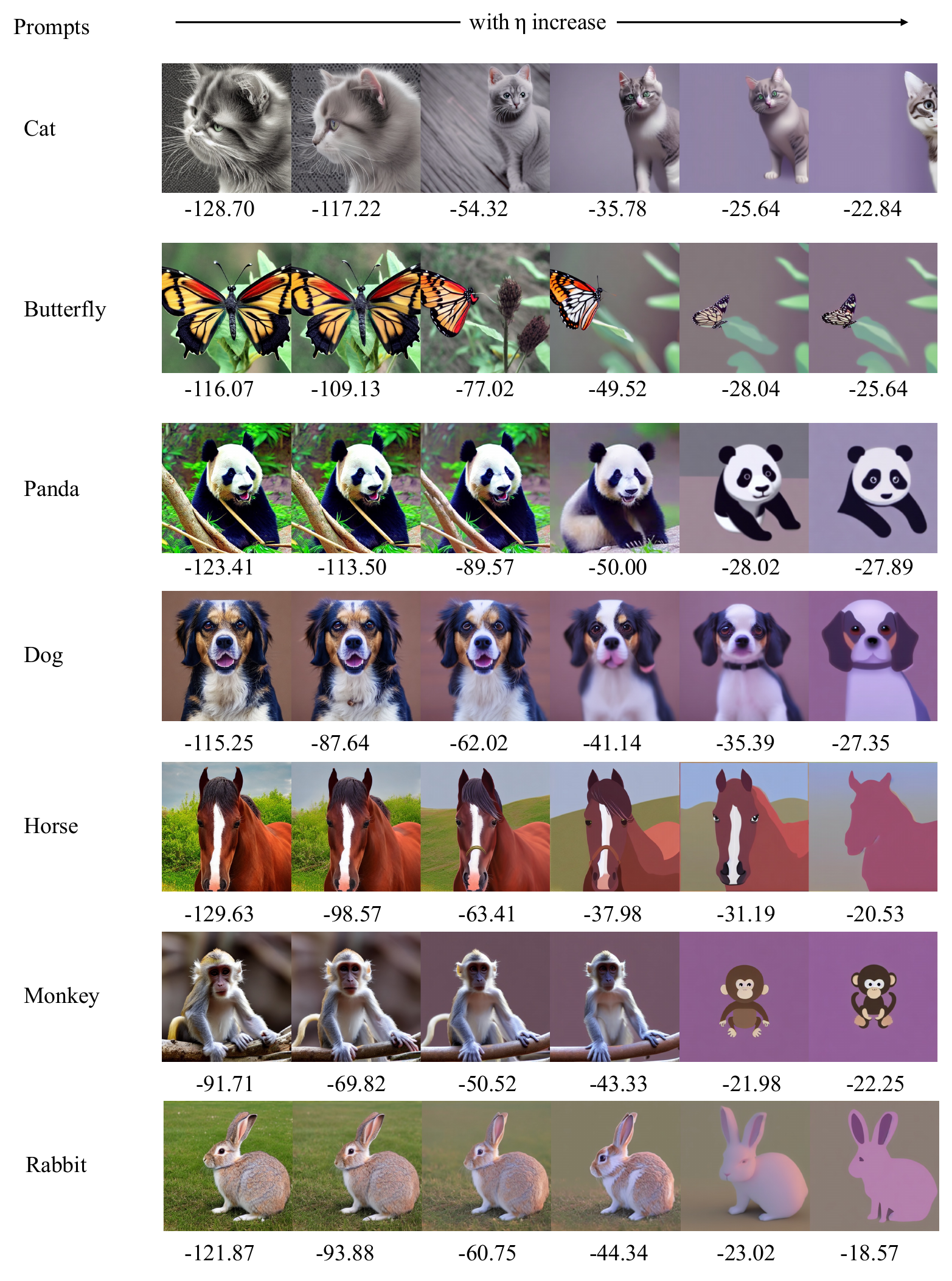}
  \vspace{-0.2in}
  \caption{More images generated by \ALG{} with varying $\eta$ values and their rewards on the image compression task.}
  \label{fig:more comp images}
\end{figure}

\cref{fig:more comp images} presents more images generated by \ALG{} on the image compression task, with different values of the parameter $\eta$. Each image is shown with its corresponding reward, demonstrating how varying $\eta$ affects the trade-off between compression efficiency and visual quality.

\begin{figure}[ht]
  \centering
  \includegraphics[width=\textwidth]{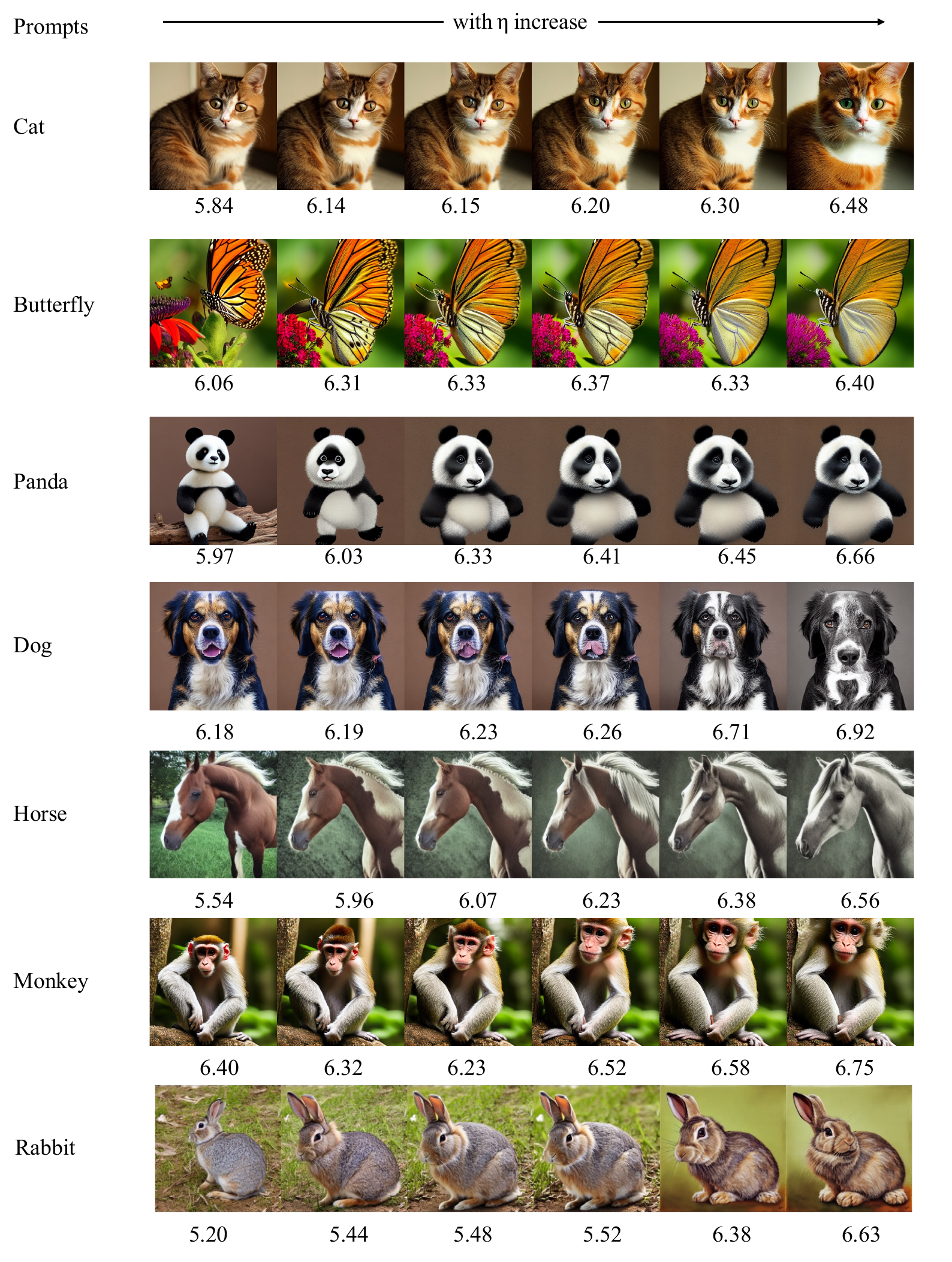}
  \vspace{-0.2in}
  \caption{More images generated by \ALG{} with varying $\eta$ values and their rewards on the image aesthetic task.}
  \label{fig:more aes images}
\end{figure}

\cref{fig:more aes images} shows additional samples from the image aesthetic task, also generated with varying $\eta$ values. These examples highlight \ALG{}'s ability to optimize for aesthetic quality under different configurations.

